\documentclass[11pt,letterpaper]{article}
\usepackage[margin=0.9in]{geometry}
\usepackage[T1]{fontenc}
\usepackage[utf8]{inputenc}
\usepackage{lmodern,microtype,amsmath,amssymb,booktabs,tabularx,xcolor,graphicx,caption,hyperref}
\usepackage[section]{placeins}
\definecolor{ink}{HTML}{243746}
\definecolor{teal}{HTML}{007E8A}
\hypersetup{colorlinks=true,linkcolor=teal,urlcolor=teal,
 pdftitle={Dungeon and Visibility World Generator},pdfauthor={}}
\newcommand{\R}{\mathbb{R}}
\newcommand{\Z}{\mathbb{Z}}
\newcommand{\W}{\mathcal{W}}
\newcommand{\code}[1]{\texttt{\small #1}}
\newcommand{\atlasfigure}[3][.37\textheight]{\par\addvspace{4pt}\noindent\begin{minipage}{\linewidth}
 \centering\includegraphics[width=\linewidth,height=#1,keepaspectratio]{#2}
 \captionof{figure}{#3}\end{minipage}\par\addvspace{4pt}}
\def\paperauthors{}
\def\paperaffiliations{}

\def\paperauthors{%
  \texorpdfstring{%
    Edwin Meriaux\textsuperscript{1,2},
    Antonio Lor\'{\i}a\textsuperscript{3},
    Gregory Dudek\textsuperscript{1}%
  }{Edwin Meriaux, Antonio Lor\'{\i}a, Gregory Dudek}%
}

\def\paperaffiliations{%
  \textsuperscript{1}McGill University,
  Montr\'eal QC H3A~0G4, Canada\\
  \textsuperscript{2}Universit\'e Paris-Saclay,
  91190 Gif-sur-Yvette, France\\
  \textsuperscript{3}CNRS, Laboratoire des Signaux et Syst\`emes (L2S),
  91190 Gif-sur-Yvette, France
}

\hypersetup{pdfauthor={\paperauthors}}
\begin{document}

{\color{ink}\LARGE\bfseries Dungeon and Visibility World Generator\par}
\vspace{5pt}
{\large Two- and three-dimensional environments for vision-based online tasks\par}
\ifx\paperauthors\empty\else
 \vspace{5pt}{\normalsize\paperauthors\par}
 {\small\paperaffiliations\par}
\fi
\section{Purpose, scope and spatial representations}
\label{sec:scope}
Robotic exploration, coverage and deployment algorithms require reproducible
environments whose geometry and visibility can be varied independently of
the algorithm. This work provides a configurable generator for two- and
three-dimensional worlds, with continuous free-space geometry, discrete
occupancy grids and geometric line-of-sight queries. It supplies benchmark
worlds and observations, rather than exploration policies.

The generator supports six geometry classes (orthogonal,
polygonal/polyhedral and curvilinear in 2D and 3D) and six construction
families (scattered rooms, clusters, winding passages, caverns, branches
and loops). Room and passage dimensions, obstacle configurations and
placement can be controlled through fixed values, uniform ranges or
probability distributions. Configurations, seeds, geometries and visibility
results are saved for reproducible comparisons.

The implementation is available at
\url{https://github.com/EdwinMeriaux/Dungeon_and_Visibility_World_Generator}.

\subsection{Continuous worlds in \texorpdfstring{$\R^2$ and $\R^3$}{R2 and R3}}
Let \(d\in\{2,3\}\). A continuous world occupies a bounded domain
\(D\subset\R^d\), with free space \(\W\subseteq D\). Positions have
real-valued coordinates: \((x,y)\) in two dimensions and \((x,y,z)\) in
three dimensions. Let \(R\) be the combined room region, \(P\) the
combined passage region, and \(O\) the combined obstacle region. Then
$ \W=\bigl(D\cap(R\cup P)\bigr)\setminus O.$

\vfill
\begin{minipage}{\linewidth}
\centering
\includegraphics[width=\linewidth,height=.28\textheight,keepaspectratio]{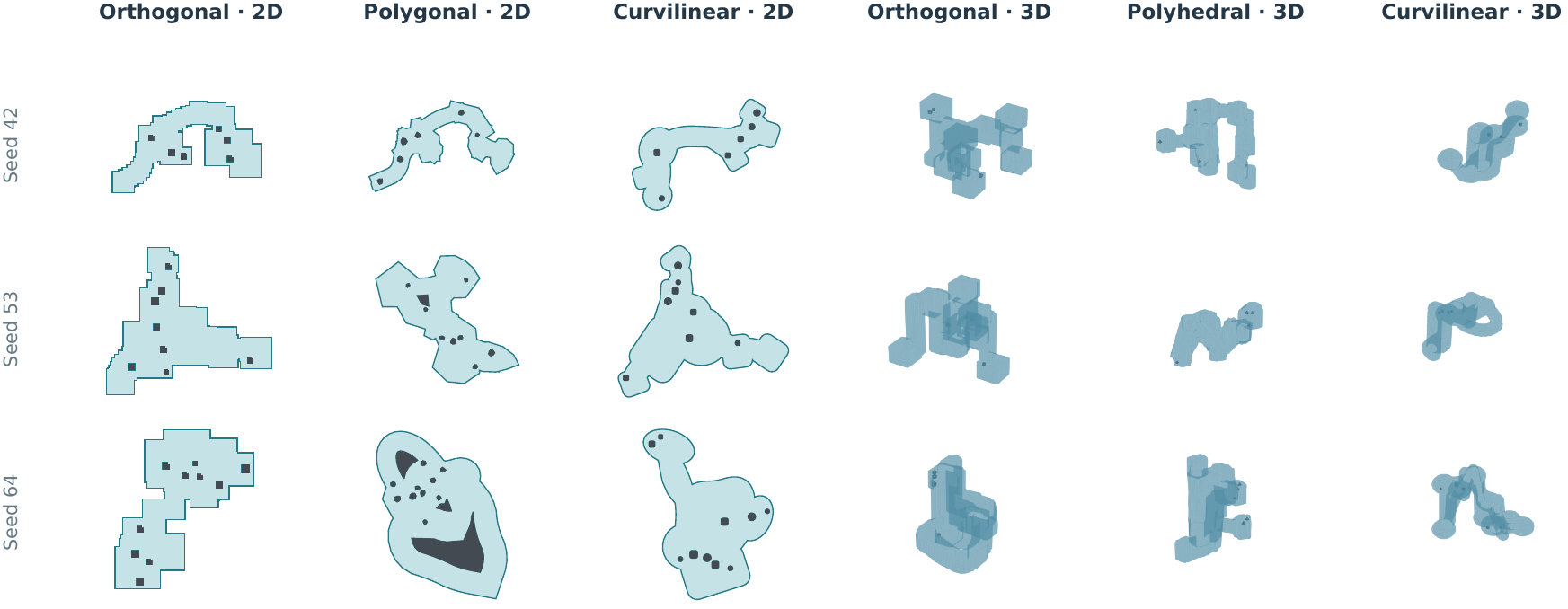}
\captionsetup{font=footnotesize}
\captionof{figure}{Eighteen generated worlds: three examples of each of the
six geometry classes. Planar free regions and 3D display surfaces give a
preview of the worlds described in Section~\ref{sec:classes}.}
\label{fig:world-mosaic}
\end{minipage}

This combines rooms and passages, keeps them inside the domain, and
removes obstacles.
The continuous representation describes wall locations without imposing a
grid. In 2D, the saved geometry is a polygon; enclosed obstacles appear as
holes, and curved boundaries are approximated by short straight segments.
In 3D, rooms and passages are combined into a solid free region, with
obstacles removed. A surface mesh is also saved for visualization, while
the solid itself is used for geometric checks.


\section{Discrete representations and visibility}
\subsection{From continuous geometry to a discrete grid}
The generator constructs \(\W\subset\R^d\), then derives a finite grid
indexed by \(\Z^d\) from the same rooms and passages.

For a 2D domain of width \(L_x\) and height \(L_y\), choose
\(n_x\) columns and \(n_y\) rows. The cell widths are
\[
 h_x=\frac{L_x}{n_x},\qquad h_y=\frac{L_y}{n_y}.
\]
With the domain starting at \((0,0)\), cell \((i,j)\) occupies the rectangle
\[
 C_{ij}=[i h_x,(i+1)h_x]\times[j h_y,(j+1)h_y].
\]
The first interval gives its horizontal extent; the second gives its
vertical extent. For unit cells, \(C_{23}=[2,3]\times[3,4]\).
In 3D, add a third interval along \(z\), with width
\(h_z=L_z/n_z\), to obtain a voxel \(C_{ijk}\).

Each cell receives a binary label:
\[
 F_{ij}=\begin{cases}
 1,&\text{if the entire cell is certified to lie in }\W,\\
 0,&\text{otherwise.}
 \end{cases}
\]
Thus the discrete free space is the set of free cell indices,
\[
 \W_{\mathrm{grid}}=\{(i,j):F_{ij}=1\}\subset\Z^2.
\]
In 3D, use \(F_{ijk}\) and triples \((i,j,k)\in\Z^3\).
The array dimensions, starting corner and cell widths are saved together.

Free labels certify whole cells. Cells cut by walls are blocked; the 3D
test can also reject wholly free voxels. On a \(50\times50\)-unit domain,
\(100\times100\) and \(40\times40\) grids have cell widths 0.5 and 1.25,
respectively. Coarser grids can thicken walls and narrow openings.

\subsection{Visibility targets and saved measurements}
Continuous targets are sampled points in \(\W\). Discrete targets are
\emph{the centers of all free cells or voxels}. A free 2D cell has center
\(t_{ij}=\bigl((i+\tfrac12)h_x,\,(j+\tfrac12)h_y\bigr)\), with \(F_{ij}=1\).
In 3D, add \((k+\tfrac12)h_z\). Let \(T=\{t_1,\ldots,t_N\}\) be the
chosen target set. For an observer \(p\), define
\[
 V(p,t)=\begin{cases}
 1,&\text{if the straight segment from }p\text{ to }t\text{ is clear},\\
 0,&\text{otherwise.}
 \end{cases}
\]
The continuous check uses the saved walls and obstacles; the discrete
check uses blocked grid cells. Boundary-contact rules are recorded with
the queries. Matched comparisons evaluate both representations at the
same free grid centers.

For a set of observers \(S\), let \(T_{\mathrm{seen}}(S)\subseteq T\)
contain the targets seen by at least one observer. The measured coverage is
\[
 \operatorname{coverage}(S)=\frac{|T_{\mathrm{seen}}(S)|}{N},
\]
where \(|T_{\mathrm{seen}}(S)|\) counts the visible targets.
Full discrete coverage means seeing every free cell or voxel center.
Neither finite target set proves visibility of every continuous point.
Coordinates, results, settings and seeds are saved for reproduction.

\section{World classes and construction families}
\label{sec:classes}
Two choices define a generated world: its \emph{geometry class} determines
the form of its walls, while its \emph{construction family} determines
how rooms and passages are arranged. A family can be realized in any of
the six geometry classes below.

\subsection{Orthogonal, polygonal and curvilinear geometry}
Orthogonal worlds have axis-aligned boundaries. General polygonal worlds
have straight edges in 2D; their 3D counterparts are polyhedral, with flat
faces. Curvilinear worlds have curved boundaries, represented by a saved
tessellation in 2D and ellipsoids and spherical passage sweeps in 3D.
\begin{center}\small
\begin{tabularx}{\linewidth}{@{}lXXX@{}}
\toprule
 & Orthogonal & Polygonal or polyhedral & Curvilinear\\
\midrule
2D & Horizontal and vertical walls & Slanted straight walls & Approximated curved walls\\
3D & Axis-aligned boxes & Flat-sided chambers and passages & Ellipsoidal chambers and rounded passages\\
\bottomrule
\end{tabularx}
\end{center}

\begin{figure}[!htbp]
\centering
\includegraphics[width=\linewidth,height=.51\textheight,keepaspectratio]{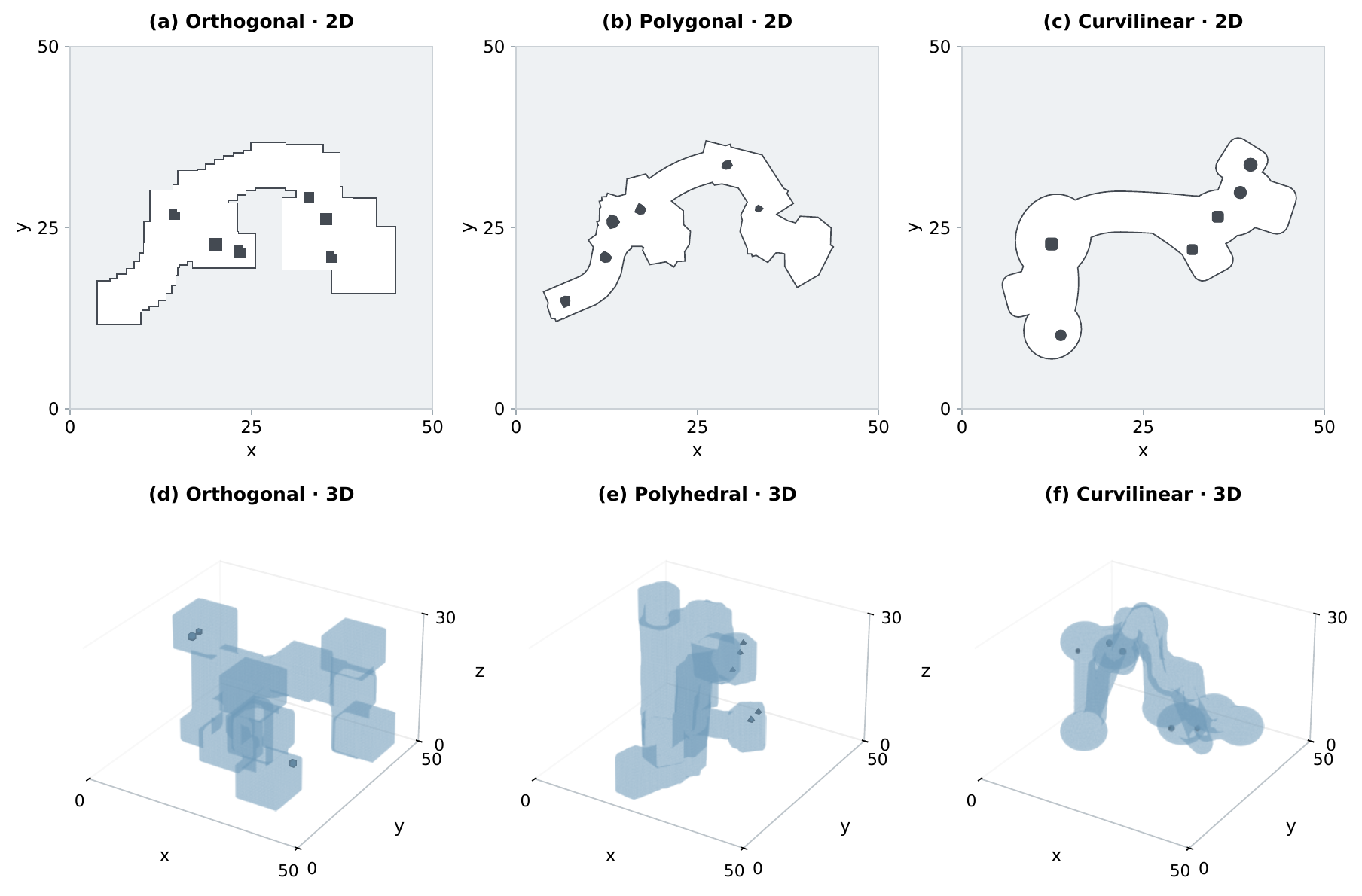}
\caption{The six geometry classes, illustrated by saved worlds.
Top: orthogonal, polygonal and curvilinear 2D worlds, each in a
\(50\times50\)-unit domain. Bottom: orthogonal, polyhedral and curvilinear
3D worlds in \(50\times50\times30\)-unit domains. White regions are
planar free space and dark islands are obstacles. Translucent blue surfaces
show 3D free-space boundaries, with dark interior obstacles. The 3D meshes
are display approximations; the saved solid defines membership and
visibility. The panels use different layouts and illustrate classes rather
than a comparison with layout held fixed.}
\label{fig:world-examples}
\end{figure}

\subsection{Six construction families}
The shared layout generator records room templates and passage paths
before class-specific walls are realized. The room-first families place
chambers and connect them; the passage-first families build a backbone
or loops and add chambers along them.

\begin{center}\small
\renewcommand{\arraystretch}{1.12}
\begin{tabularx}{\linewidth}{@{}lX@{}}
\toprule
Family & Construction before geometric realization\\
\midrule
Scattered rooms & Uneven polar placement of unequal chambers, followed by graph-connected bent paths.\\
Room clusters & Random anchors and Gaussian groups of room centers, joined through candidate connections.\\
Caverns & A small population of large central chambers connected through graph edges.\\
Winding passages & A smoothed main passage with selected chambers and angled side branches.\\
Branches & A central chamber with turning arms and rooms along their growth.\\
Irregular loops & Deformed closed passage loops, attached chambers and links between loops.\\
\bottomrule
\end{tabularx}
\end{center}

\begin{figure}[!htbp]
\centering
\includegraphics[width=\linewidth,height=.48\textheight,keepaspectratio]{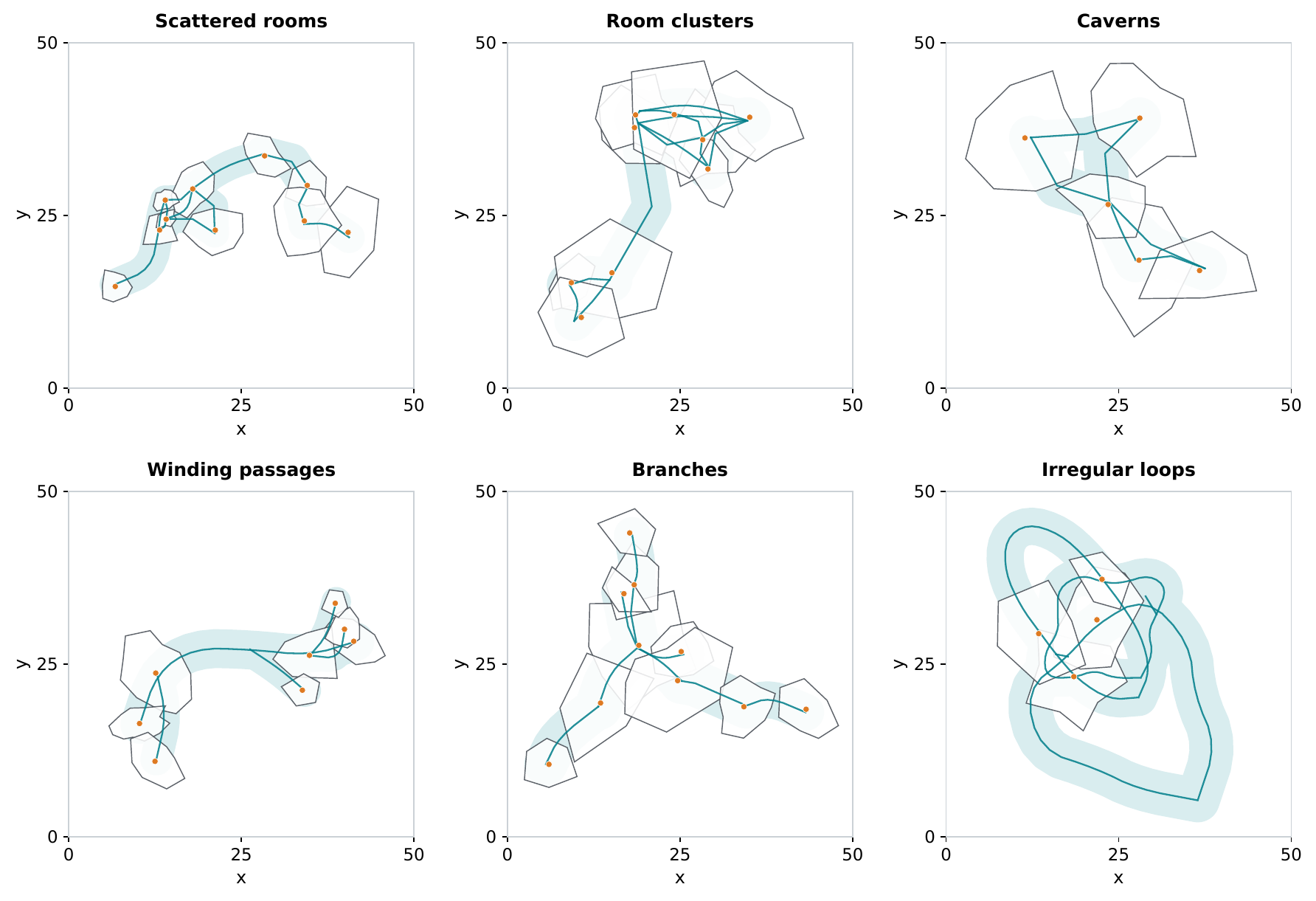}
\caption{One saved construction layout for each family, before
class-specific geometric realization. White outlined shapes are room
templates, orange points are their saved centers, teal lines are passage
centerlines, and pale teal bands show the recorded passage widths.
These are construction elements before union; overlapping rooms or paths
can merge in the final world. Inserted obstacles are omitted. All panels
use a \(50\times50\)-unit planar domain. The 3D realization assigns
elevations and adds inter-elevation connections to the planar layout.}
\label{fig:construction-families}
\end{figure}

These descriptions give the default placement patterns. Configured spatial
distributions can replace or perturb room positions; size distributions
control their dimensions. The saved layout retains the realized room
centers and passage paths so that the construction can be inspected.

\section{How the code generates and records a world}
\label{sec:code}
The entry point is \path{scripts/run_pipeline.py}. Python constructs the
world, derives its occupancy grid and selects sample points. A standalone
Rust program computes visibility. The workflow is
\[
 \text{configuration}\;\longrightarrow\;\text{layout}\;\longrightarrow\;
 \text{continuous world}\;\longrightarrow\;\text{grid and queries}.
\]
\subsection{Read the configuration and sample parameters}
The JSON configuration specifies the dimension, geometry class,
construction family, world size, grid resolution and random seed.
\path{python/wgv/parameters.py} interprets a scalar as a fixed value,
a two-value range as a uniform law, or a named distribution as a sampling
rule. This fragment specifies uniform room positions and dimensions:
\begin{quote}\small
\begin{verbatim}
"rooms": {"count": [7, 10],
          "width": [6, 9], "length": [7, 11],
          "placement": {"type": "independent",
                        "x": [12, 38], "y": [12, 38]}}
\end{verbatim}
\end{quote}
The room count is drawn from the integers 7 through 10. For each room,
\(w_i\sim\mathcal U(6,9)\), and its two position coordinates are
independently drawn from \(\mathcal U(12,38)\). Joint Gaussian and mixture
laws can express correlated or clustered placement. World settings are
drawn per candidate; dimensions are drawn per room, path or obstacle.

\subsubsection*{Example: Gaussian room placement}
Keep the same uniform count and size ranges, and change only the placement:
\begin{quote}\small
\begin{verbatim}
"rooms": {"count": [7, 10],
          "width": [6, 9], "length": [7, 11],
          "placement": {"type": "gaussian",
                        "mean": [25, 25],
                        "std": [6, 6]}}
\end{verbatim}
\end{quote}
For each room center \((x_i,y_i)\), this gives independent coordinates
\[
 x_i\sim\mathcal N(25,6^2),\qquad y_i\sim\mathcal N(25,6^2).
\]
Room centers concentrate around \((25,25)\), with standard deviation 6
world units along each axis. The Gaussian has no hard coordinate bounds;
world fitting and acceptance still apply. A \texttt{covariance} matrix can
replace \texttt{std} to specify correlated coordinates.

Room counts, dimensions and placement, as well as the exposed passage and
obstacle parameters, can each be controlled through fixed values, uniform
ranges or other supported distributions. Different parameters can use
different laws; one distribution need not govern the entire world.
The following tables list the implemented choices. Integer controls and
the selected construction family determine which specifications are valid.

\subsubsection*{Supported scalar distributions}
The following names are implemented in \path{python/wgv/parameters.py}.
They apply to scalar controls supported by the selected construction family.
Fixed values and uniform ranges are also available as shorthand.
\begin{center}\small
\renewcommand{\arraystretch}{1.18}
\begin{tabularx}{\linewidth}{@{}p{.25\linewidth}X@{}}
\toprule
\texttt{distribution} & Parameters and meaning\\
\midrule
\texttt{fixed} & \texttt{value}: always return this value. A bare number is equivalent.\\
\texttt{uniform} & \texttt{min}, \texttt{max}: uniform real-valued draw over the range.\\
\texttt{uniform\_int} & \texttt{min}, \texttt{max}: uniform integer draw, including both endpoints.\\
\texttt{normal},\newline\texttt{gaussian} & \texttt{mean}, \texttt{std}: two names for the same Gaussian law. Integer controls round; optional integer bounds reject rounded values outside the range.\\
\texttt{truncated\_normal} & \texttt{mean}, \texttt{std}, \texttt{min}, \texttt{max}: condition a Gaussian draw on the range before any integer rounding.\\
\texttt{lognormal} & \texttt{mean} (default 0), \texttt{sigma}: mean and standard deviation of the underlying normal variable, in log space.\\
\texttt{exponential} & \texttt{scale}: a nonnegative exponential draw with this positive mean.\\
\texttt{poisson} & \texttt{lambda}: a nonnegative integer count with this mean.\\
\texttt{beta} & \texttt{alpha}, \texttt{beta}: positive shape parameters. Optional \texttt{min}, \texttt{max} rescale the default interval \([0,1]\).\\
\texttt{categorical} & \texttt{values}, optional \texttt{weights}: choose one listed value; equal probabilities if weights are omitted.\\
\texttt{mixture} & \texttt{components}, optional \texttt{weights}: choose a component, then sample its law; equal component probabilities by default.\\
\bottomrule
\end{tabularx}
\end{center}
The shorthand \texttt{[min,max]} selects \texttt{uniform\_int} for an
integer field and \texttt{uniform} otherwise. Continuous Gaussian values
require \texttt{truncated\_normal} to enforce bounds. Gaussian integer
bounds must themselves be integers. Other laws do not automatically round
fractional draws for integer fields. Categorical and mixture weights need
not sum to one: the code normalizes them.

\subsubsection*{Supported spatial placement rules}
Locations use a \texttt{type}, with one coordinate per spatial dimension.
\begin{center}\small
\renewcommand{\arraystretch}{1.16}
\begin{tabularx}{\linewidth}{@{}p{.25\linewidth}X@{}}
\toprule
\texttt{type} & Parameters and meaning\\
\midrule
\texttt{fixed} & \texttt{value}: a prescribed coordinate vector.\\
\texttt{independent} & \texttt{x}, \texttt{y}, and \texttt{z} in 3D: independently sample each axis using any scalar law above.\\
\texttt{uniform} & Vector \texttt{min}, \texttt{max}: uniform placement within a box.\\
\texttt{gaussian},\newline\texttt{normal} & Vector \texttt{mean} and scalar/per-axis \texttt{std}, or a full \texttt{covariance} matrix for correlated coordinates.\\
\texttt{mixture} & Spatial \texttt{components}, optional \texttt{weights}: mix placement rules, for example Gaussian clusters.\\
\bottomrule
\end{tabularx}
\end{center}
The layout also offers \texttt{family} placement with sampled offsets from
family-generated centers; this is a construction mode rather than another
probability law. Whole-world acceptance still applies to sampled geometry
(Section~\ref{sec:acceptance}).


\begin{figure}[!htbp]
\centering
\includegraphics[width=\linewidth]{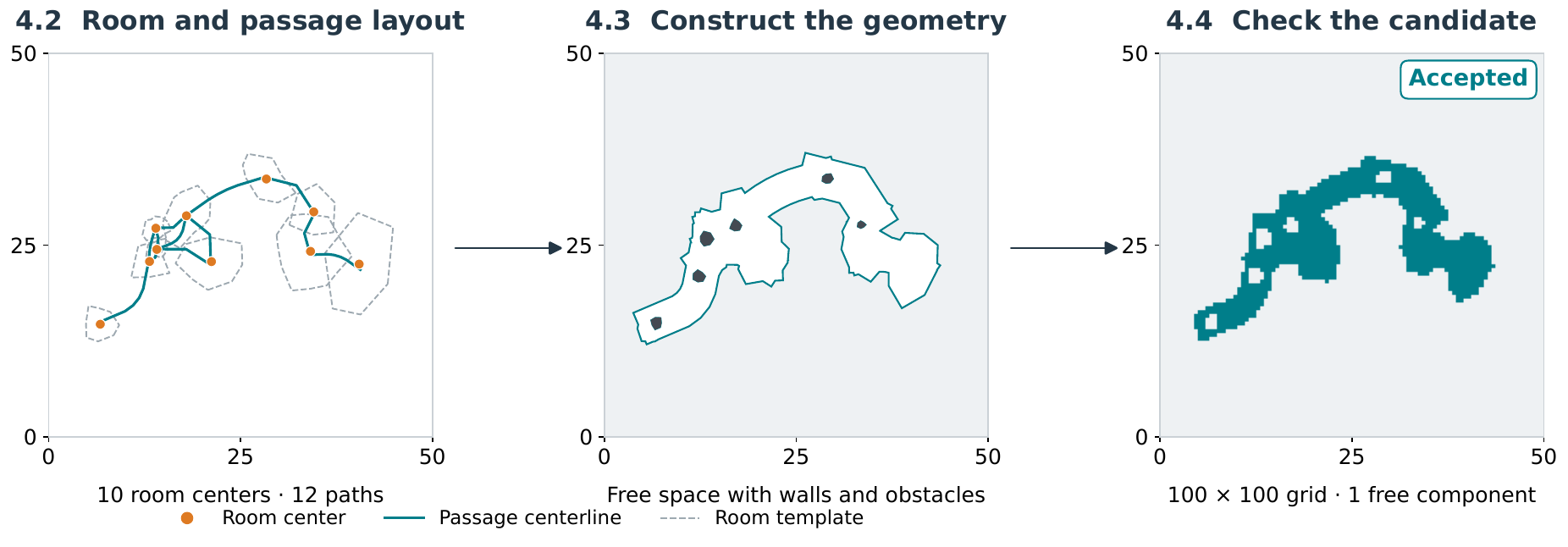}
\caption{Three stages for the same saved 2D world. Left: room templates,
centers and passage paths. Middle: the constructed polygon with finite-width
passages and blocked holes. Right: its conservative grid has one connected
free component. Checking accepts or rejects a candidate; it does not alter
its geometry.}
\label{fig:generation-stages}
\end{figure}

\subsection{Choose the room and passage layout}
\label{sec:layout-stage}
First decide \emph{where rooms go and which passages connect them}
(Figure~\ref{fig:generation-stages}, left). \path{python/wgv/layout.py}
produces room centers, templates and passage centerlines. Room-first
families connect rooms through a spanning tree and extra edges; passage-first
families grow a backbone, branches or loops. Explicit distributions use
\path{python/wgv/layout_distributions.py}. A \texttt{layout\_file} can
reuse a saved layout. Requested and achieved room counts are recorded.

\subsection{Construct the continuous geometry}
\label{sec:geometry-stage}
Next turn that layout into \emph{actual free space with walls}
(Figure~\ref{fig:generation-stages}, middle).
\path{python/wgv/generation.py} creates rooms in the selected geometry
class, gives passages finite width, combines their regions and subtracts
obstacles. In 2D, \texttt{realize\_2d()} uses orthogonal, polygonal or
tessellated curved boundaries. In 3D, \texttt{realize\_3d()} uses boxes,
polyhedral prisms or ellipsoids, with box, prism or overlapping-ball
passages. Fitting is recorded; \texttt{fit\_mode="none"} preserves
explicit positions and rejects geometry outside the domain.

\subsection{Check the candidate and accept or regenerate}
\label{sec:acceptance}
Finally decide \emph{whether the candidate meets the requirements}
(Figure~\ref{fig:generation-stages}, right). The checks require valid,
connected continuous free space, safe obstacle placement and the configured
grid connectivity. In 3D, primitive-overlap certificates establish connections.
The grid is derived by Section~4.5's rule; narrow passages may disappear
and cause rejection.

A passing candidate is saved. Otherwise \texttt{generate()} retries with
a new attempt seed, records the reason and raises an error at
\texttt{max\_attempts}. Checking does not reshape a candidate.
Acceptance conditions the requested distributions; saved attempts and
resolved settings document this selection.

\subsection{Derive the occupancy array}
\begin{figure}[!htbp]
\centering
\includegraphics[width=\linewidth]{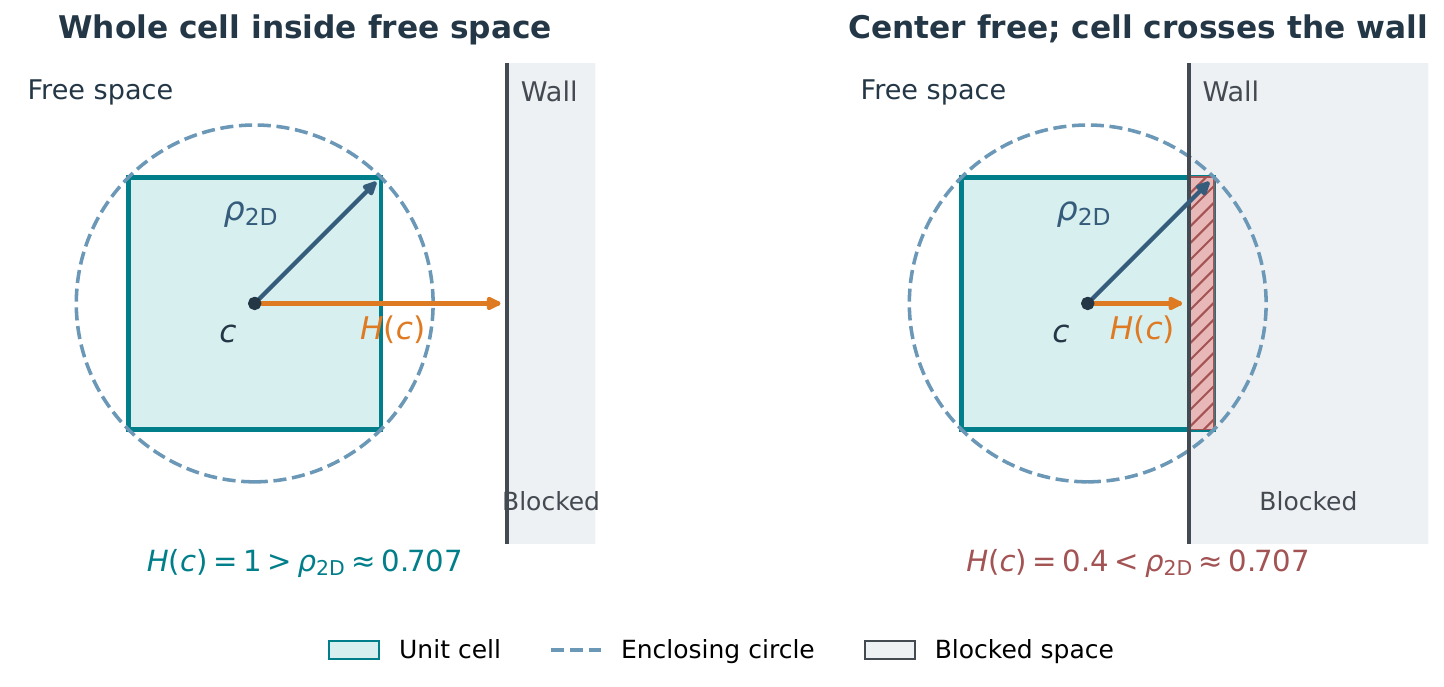}
\caption{A 2D illustration of whole-cell safety. Left: the entire unit
square lies in free space; \(H(c)=1>\rho_{\mathrm{2D}}\). Right: its center
is free, but part of the square crosses the wall. The enclosing circle
illustrates the clearance principle used with a sphere in 3D; the actual
2D implementation tests whole-square containment directly.}
\label{fig:cell-clearance-2d}
\end{figure}

\texttt{discrete()} derives a dense Boolean array as in Section~2.1.
Shapely tests whole-cell containment in 2D. In 3D, the code evaluates
\(H(c)\), a guaranteed lower bound on the clearance from voxel center
\(c\) to blocked space. With voxel widths \(h_x,h_y,h_z\), define
\[
 \rho=\tfrac12\sqrt{h_x^2+h_y^2+h_z^2}.
\]
This is the distance from the center to the farthest corner. A sphere of
radius \(\rho\) around \(c\) encloses the entire voxel. The code labels
the voxel free only when
\[
 H(c)>\rho+10^{-9}.
\]
In words, the guaranteed center clearance must exceed the center-to-corner
distance, with a tiny numerical safety margin. This certifies the whole
voxel, rather than just its center. Figure~\ref{fig:cell-clearance-2d}
illustrates the same idea with a 2D square, whose center-to-corner distance
is \(\rho_{\mathrm{2D}}=\sqrt{2}/2\approx0.707\). For a unit 3D cube,
\(\rho=\sqrt{3}/2\approx0.866\). The 2D code checks square containment
directly; the 3D code uses the clearance certificate above.
Failing the 3D check gives a blocked label, even though some rejected voxels
may be wholly free. Connectivity uses four neighbors in the 2D showcases
and six in 3D.

\subsection{Choose continuous samples or discrete grid centers}
\begin{figure}[!htbp]
\centering
\includegraphics[width=\linewidth]{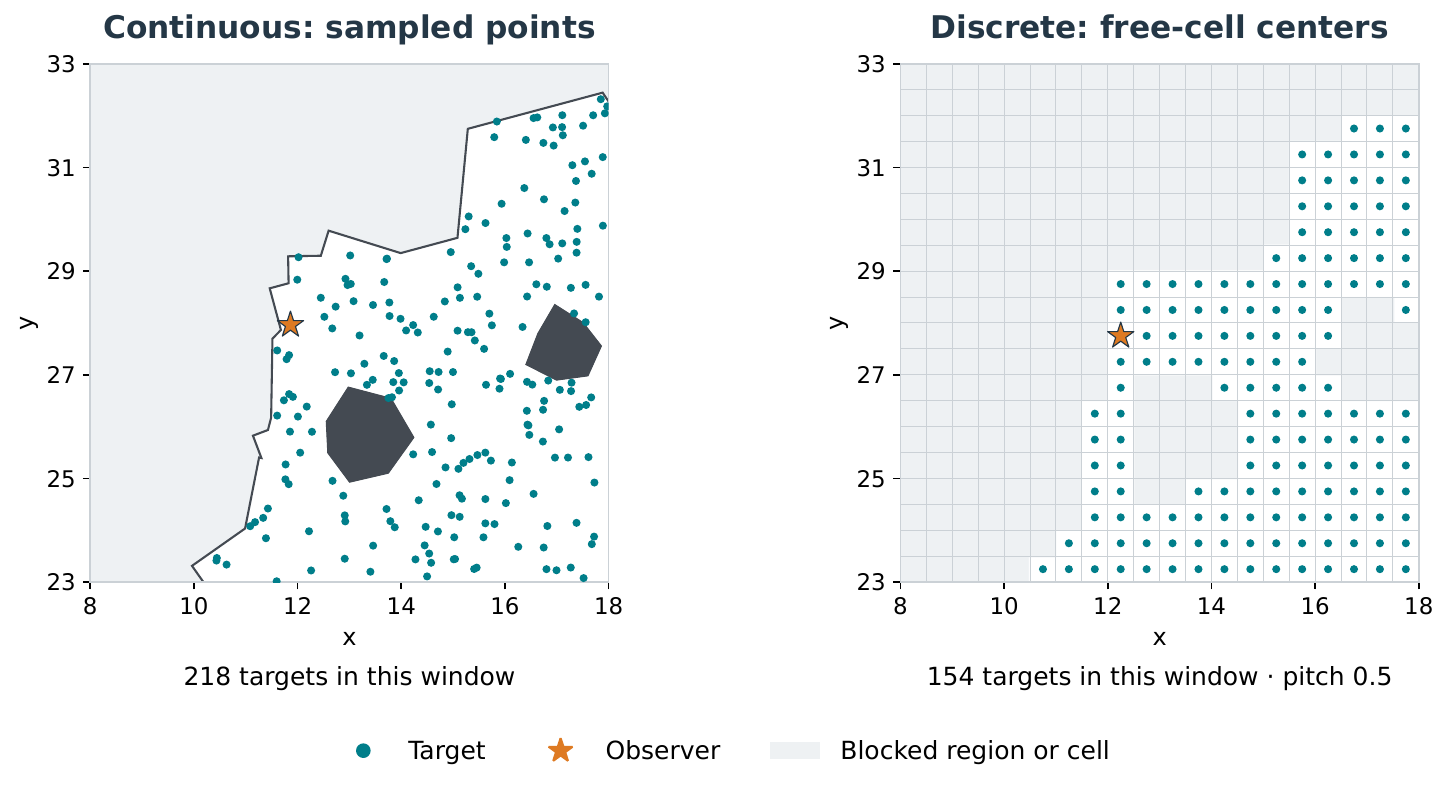}
\caption{The same window of a saved 2D world. Left: continuous sample
points. Right: one target per free cell, with spacing 0.5. Orange stars
show one continuous observer and its assigned free grid center.}
\label{fig:continuous-grid-targets}
\end{figure}

\path{python/wgv/sampling.py} prepares two target sets
(Figure~\ref{fig:continuous-grid-targets}). For continuous visibility,
\texttt{prepare()} defaults to 1500 uniform interior targets and eight
observers. Sampling uses a separate seed, independently of grid resolution;
observers follow a deterministic farthest-point sequence on the targets.

For discrete visibility, \texttt{prepare\_discrete()} uses \emph{one target
at the center of every free cell or voxel}. Its target count equals the
number of free grid entries. Saved origin and spacing convert indices to
world coordinates. Each continuous observer is assigned its nearest
distinct free grid center. Matched queries use these same center coordinates.

Continuous sampling also supports \texttt{gaussian}, \texttt{clearance},
\texttt{poisson\_disk}, \texttt{lattice}, \texttt{boundary} (2D),
\texttt{explicit} and a Python \texttt{callback}. Discrete targets come
from the occupancy grid.

\subsection{Compute the visibility matrix}
\path{python/wgv/visibility.py} saves the world, observer and target
coordinates, and sensor settings in a JSON query, then invokes
\path{rust/visibility_engine}. The output represents
\[
 A_{ij}=V(p_i,t_j),\qquad 1\leq i\leq M,\quad 1\leq j\leq N.
\]
Each row lists one observer's visible targets. In 2D, Rust checks polygon
boundary edges. In 3D, intervals inside positive primitives must cover
the segment, with no negative-obstacle contact under the recorded margins.
Discrete queries use supercover DDA, checking side cells at edge and
corner crossings. Sensing defaults to unlimited range and no directional
restriction; configured sensor rules can reject additional targets.

\subsection{Save, validate and replay the results}
The matrix uses compressed sparse row (CSR) form: each observer lists its
visible target indices. Python also saves a SciPy matrix, visible-union
indices and unobserved indices for the coverage in Section~2.2.

\texttt{paired\_representations()} compares both representations at the same
grid centers and records observer shifts and visibility agreement. Earlier
bundled results use continuous samples inside free cells, without snapping;
they remain historical evidence, replayable with \texttt{legacy\_common\_samples}.

\begin{center}\small
\renewcommand{\arraystretch}{1.12}
\begin{tabularx}{\linewidth}{@{}p{.43\linewidth}X@{}}
\toprule
Saved artifact & Contents\\
\midrule
\path{configuration.json},\newline\path{resolved_configuration.json} & Requested settings and settings resolved during the accepted attempt.\\
\path{layout.json}, \path{metadata.json} & Templates, paths, realized parameters, counts, seeds, attempt history and source hashes.\\
\path{geometry.geojson} or \path{solid.json} & The continuous polygon or constructive solid.\\
\path{free.npy} & Conservative occupancy; origin, spacing and axis order are in metadata.\\
\path{targets.npy}, \path{observers.npy},\newline\path{sampling.json} & Continuous samples, sampling seed and observer selection.\\
\path{discrete_targets.npy},\newline\path{discrete_observers.npy},\newline\path{discrete_sampling.json} & All free grid centers and adjusted observers; integer indices are saved in \path{discrete_target_cell_indices.npy}.\\
\path{visibility_continuous_input.json} & The query passed to Rust, including geometry and numerical settings.\\
\path{visibility_continuous.csr},\newline\path{visibility_continuous.npz},\newline\path{visibility_continuous.json} & Visibility rows and summaries; paired queries have separate labeled files.\\
\path{connectivity_certificate.json},\newline\path{validation.json} & Connection evidence and consistency checks.\\
\bottomrule
\end{tabularx}
\end{center}

\path{python/wgv/validation.py} checks geometry hashes, dimensions,
connectivity, target coordinates and visibility outputs. It independently
reconstructs 2D occupancy by cell-minus-polygon subtraction and checks
3D voxel corners and connection witnesses. \path{python/wgv/visualization.py}
renders results; display meshes do not replace the visibility solid.

The driver's \texttt{all} stage follows this sequence, with the retry
loop inside \texttt{generate()}:
\begin{quote}\small
\begin{verbatim}
generate(config, folder)
prepare(folder, count=1500, observers=8, seed=7001)
compute(folder, mode="continuous")
compute(folder, mode="discrete")
paired_representations(folder)
validate(folder)
render_world(folder, figure_stem)
\end{verbatim}
\end{quote}
Stages run separately; matching saved worlds are preserved. Changed
configurations require fresh output folders. Reproduction requires the
same source and dependency versions. Saved query inputs allow Rust replay
without regeneration or resampling.


\section{Generation time and reproducibility}
\label{sec:generation-reproducibility}
This section measures the cost of generating representative worlds and
checks whether the same configuration and seed reproduce their saved
geometry and occupancy. The experiment uses one configuration from each
of the six classes in Section~\ref{sec:classes}. These are six distinct
worlds; repeated runs reconstruct each world rather than draw new seeds.

\subsection{Measured generation time}
For each configuration, the study performs one excluded warm-up followed
by five timed calls to \texttt{generate(config, output)}. Every call writes
to a new directory. The runs execute serially on an Apple M2 Max with
12 CPU cores and 64\,GiB of memory, using macOS~26.3, Python~3.11.15,
NumPy~1.26.4, SciPy~1.16.3, Shapely~2.1.2 and GEOS~3.13.1.
BLAS and OpenMP thread limits are set to one.

The measured wall-clock interval begins immediately before the generation
call and ends when it returns. It includes parameter draws, layout
construction, continuous geometric realization, obstacle insertion,
candidate rejection, occupancy conversion and all generation-artifact
writes. It excludes Python imports, configuration-file reads, the
warm-up, hash comparisons, target selection, visibility computation and
display-mesh rendering. Thus the table reports the cost of creating and
saving a world, rather than running the complete sensing pipeline.

All configurations use world seed 42 and request ten room templates.
The 2D domains measure \(50\times50\) world units, with
\(100\times100=10{,}000\) grid entries. The 3D domains measure
\(50\times50\times30\) units, with
\(70\times70\times42=205{,}800\) voxel entries. Table~\ref{tab:generation-times}
also records the achieved room and passage counts and the number of free
entries; occupancy conversion evaluates the entire array.

\begin{table}[!htbp]
\centering\small
\begin{tabular}{@{}llrrrr@{}}
\toprule
Class & Family & Rooms/passages & Free entries & Median (s) & Range (s)\\
\midrule
2D Orthogonal & Scattered rooms & 10/12 & 1,775 & 0.027 & 0.026--0.033\\
2D Polygonal & Scattered rooms & 10/12 & 1,127 & 0.025 & 0.025--0.032\\
2D Curvilinear & Winding passages & 8/5 & 1,503 & 0.029 & 0.028--0.030\\
\midrule
3D Orthogonal & Room clusters & 9/10 & 18,512 & 1.909 & 1.888--1.950\\
3D Polyhedral & Scattered rooms & 10/12 & 8,179 & 2.853 & 2.781--3.308\\
3D Curvilinear & Branches & 9/8 & 8,409 & 3.162 & 3.144--3.344\\
\bottomrule
\end{tabular}
\caption{Complete generation-call times for one fixed world per class.
Medians and minimum--maximum ranges use five timed repeats after one
excluded warm-up per configuration. Rooms/passages count the realized
construction elements before their union. All configurations use seed 42.
Full timings and configurations are archived with the study.}
\label{tab:generation-times}
\end{table}


Median times are approximately 0.03 seconds in 2D and 1.9--3.2 seconds in
3D. The curved 3D case needs two candidate attempts; the other five cases
succeed on their first. Attempt histories are identical across repeats,
and the timings include rejected candidates.

These measurements describe six configurations on one host. Layouts,
primitive counts and obstacles differ, so they do not isolate the effect
of wall class or establish a scaling law. The repeats describe runtime
variation for fixed worlds; other seeds, resolutions and rejection
histories can change the cost. Runtime is distinct from spatial
complexity and sensing difficulty.

\subsection{Exact reproduction of the saved world}
After the five timed runs, each configuration is generated once more in
a fresh Python process. Across these 36 saved outputs, each world's
continuous geometry, occupancy array and construction records match its
first timed run byte for byte. SHA-256 hashes are calculated for every
generation artifact except \path{metadata.json}; the metadata is compared
as a JSON object after removing only \texttt{generation\_seconds}.
Directory names and measured times can differ while the generated world
remains identical.

\begin{table}[!htbp]
\centering\small
\begin{tabularx}{\linewidth}{@{}Xcc@{}}
\toprule
Compared artifact & Timed comparisons & Fresh-process replay\\
\midrule
\path{geometry.geojson} or \path{solid.json} & 24/24 match & 6/6 match\\
\path{free.npy} & 24/24 match & 6/6 match\\
\path{layout.json} & 24/24 match & 6/6 match\\
Requested and resolved configurations & 24/24 match & 6/6 match\\
\path{connectivity_certificate.json} & 24/24 match & 6/6 match\\
\path{primitives.geojson} (2D only) & 12/12 match & 3/3 match\\
Metadata, excluding generation time & 24/24 match & 6/6 match\\
\bottomrule
\end{tabularx}
\caption{Reproduction checks against each world's first timed run.
There are six baselines, 24 additional timed comparisons and six
fresh-process comparisons. No geometric or
coordinate tolerance is used to declare artifact equality.}
\label{tab:generation-reproduction}
\end{table}

The generator creates a new random stream from the world seed and
candidate-attempt index. Repeating the full configuration with the same
code and dependency versions therefore repeats parameter draws and
candidate decisions in these examples. A reproducibility record must
retain the complete requested configuration, source hashes and software
versions, together with any referenced frozen layout or external inputs.
The tests establish exact reproduction in the recorded environment;
they do not establish byte identity across dependency upgrades or
different computing platforms.

World generation and continuous target sampling use separate seeds.
To reproduce continuous target coordinates, also retain the sampling
method, target count and sampling seed. Discrete targets are derived
from the free-cell or free-voxel centers and require no additional
target-sampling seed. Observer selection and visibility reproduction
also require their saved settings or query inputs
(Section~\ref{sec:code}). These operations are outside the generation
timings and artifact comparisons above.

\subsection{Saved evidence and repeating the experiment}
The study saves its evidence in \path{datasets/}, under
\texttt{generation\_reproducibility}. This folder contains the six input
configurations, all 36 generated outputs, individual timings in
\path{timings.csv}, and full hashes, source identities, software versions
and summary statistics in \path{report.json}. The six configurations
are the \texttt{\_1.json} examples in \path{configs/showcase/}.

From the project root, rerun the study into a new directory:
\begin{quote}\small
\begin{verbatim}
.venv/bin/python scripts/generation_reproducibility_study.py \
  --output datasets/my_generation_study --repeats 5
\end{verbatim}
\end{quote}
The script rejects a populated output directory and stops if any
reproduction comparison fails. The archived worlds remain available
for independent inspection even when rerunning the timing experiment
produces different elapsed times.



\section{World examples and saved visibility}
\label{sec:atlas}
The earlier atlas is retained here with its saved geometry and sensing
results. It contains three examples of each class, at seeds 42, 53 and 64.
All planar domains span \(50\times50\) units; volumetric domains span
\(50\times50\times30\). Their occupancy arrays contain
\(100\times100\) cells or \(70\times70\times42\) voxels.

Every example uses 1500 continuous interior targets and eight illustrative
observers. Orange marks observers, teal marks visible targets and gray
marks occluded targets. The occupancy panels show the derived grids;
the displayed continuous targets are separate from the free grid centers
used by the current discrete-query convention (Section~4.6).
Target fractions are measured on the saved continuous point sets.

All showcased continuous worlds and free grids are connected. Rooms and
passages are construction counts before union. Planar visibility shading
and 3D display meshes are numerical illustrations; Rust determines the
saved target classifications from the polygon or solid.

\FloatBarrier
\subsection{Orthogonal worlds in two dimensions}

\atlasfigure[.33\textheight]{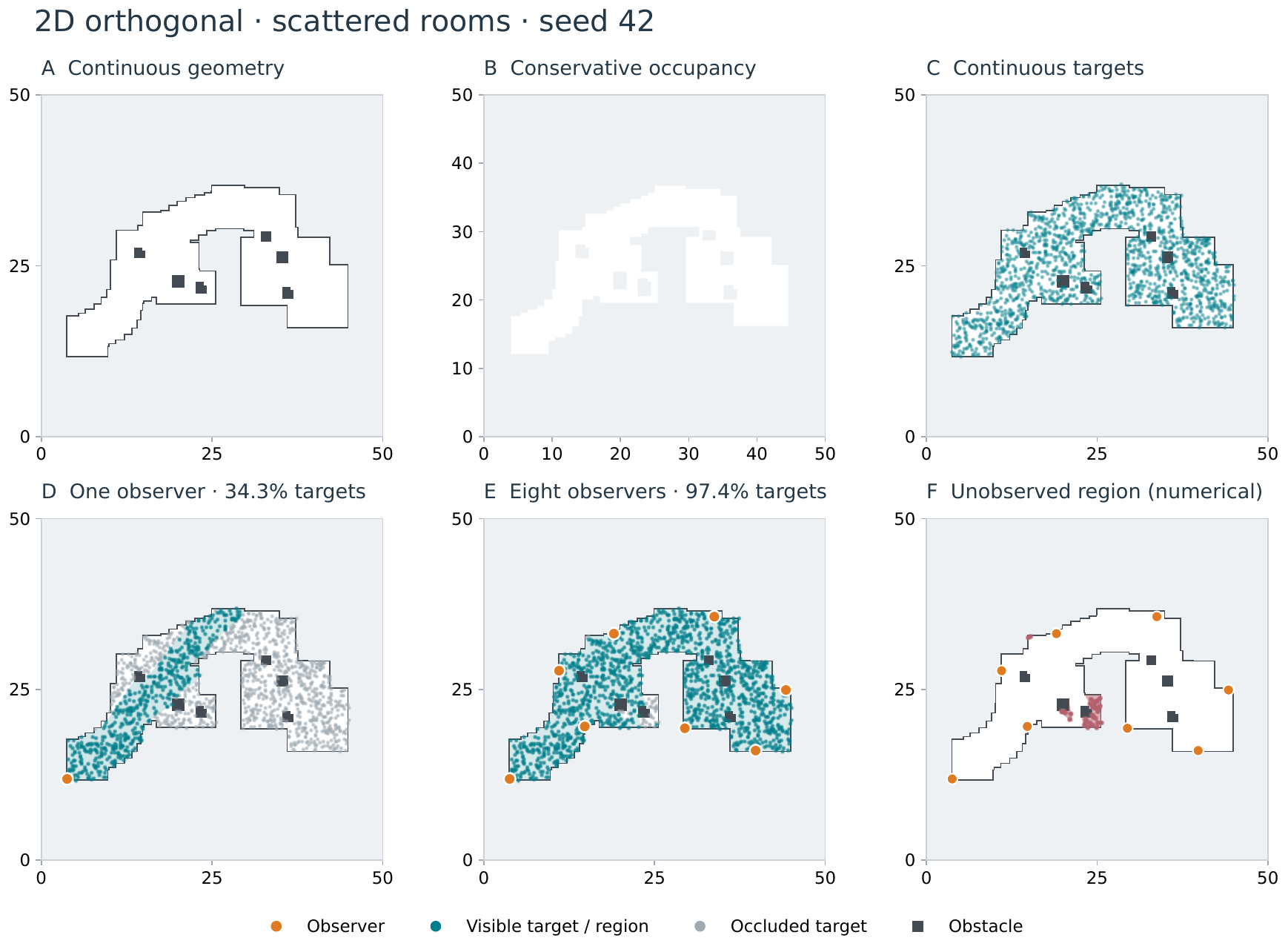}{2D orthogonal, $50\times50$ world units, scattered rooms, seed 42: 10 realized rooms, 12 original layout paths, 6 inserted obstacles, graph cycle rank 3, and 6 polygon holes. Occupancy uses $100\times100$ cells. Free area 497.2 square units. The 1,500 interior targets and eight observers give 97.4\% union target coverage using polygon-edge visibility. }

Two more seeds change the family and realized world.

\atlasfigure[.33\textheight]{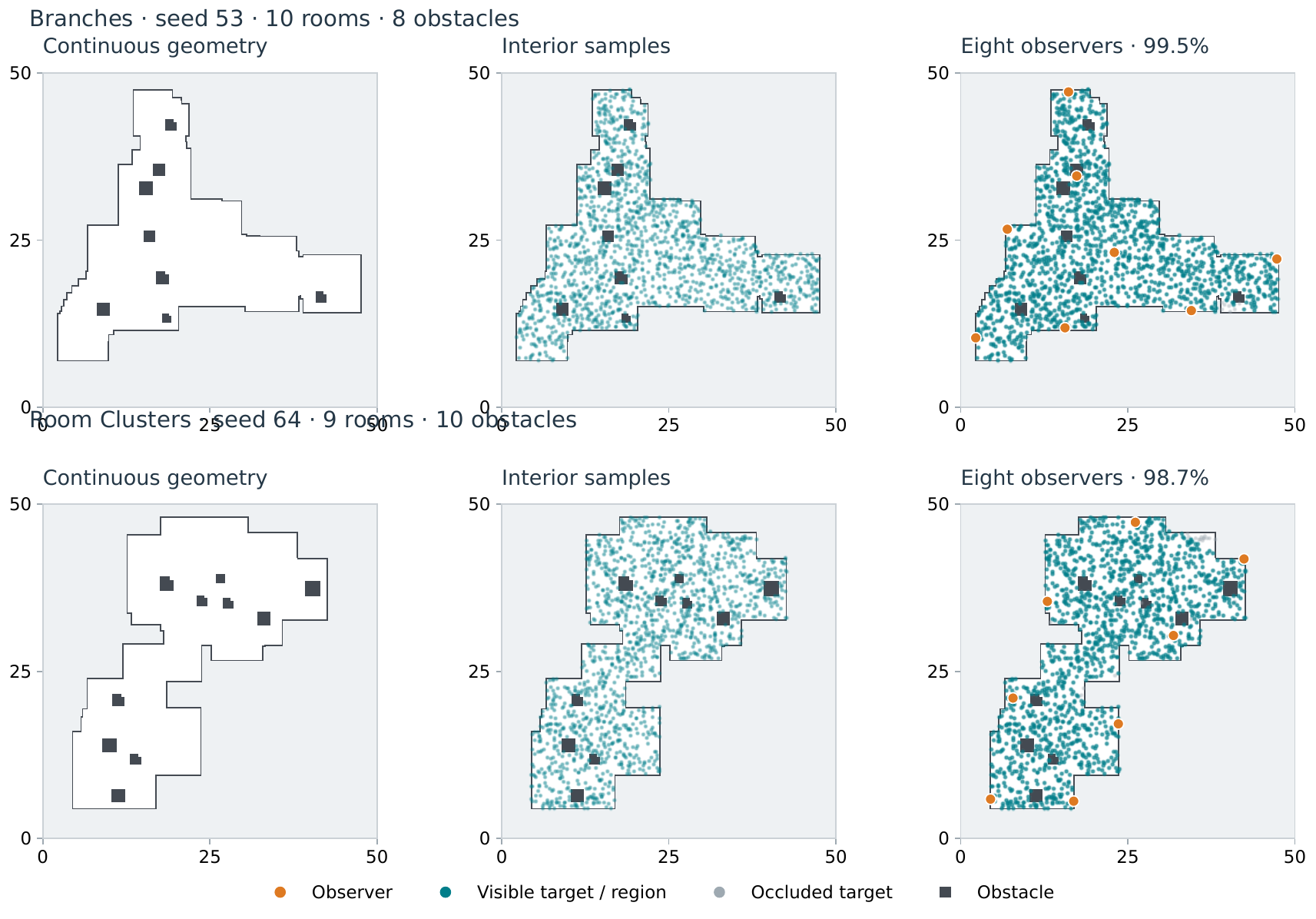}{2D orthogonal, $50\times50$ world units, branches, seed 53: 10 realized rooms, 9 original layout paths, 8 inserted obstacles, graph cycle rank 0, and 8 polygon holes. Occupancy uses $100\times100$ cells. Free area 760.6 square units. The 1,500 interior targets and eight observers give 99.5\% union target coverage using polygon-edge visibility.  2D orthogonal, $50\times50$ world units, room clusters, seed 64: 9 realized rooms, 15 original layout paths, 10 inserted obstacles, graph cycle rank 7, and 10 polygon holes. Occupancy uses $100\times100$ cells. Free area 830.3 square units. The 1,500 interior targets and eight observers give 98.7\% union target coverage using polygon-edge visibility. }

\FloatBarrier
\subsection{Polygonal worlds in two dimensions}

\atlasfigure[.33\textheight]{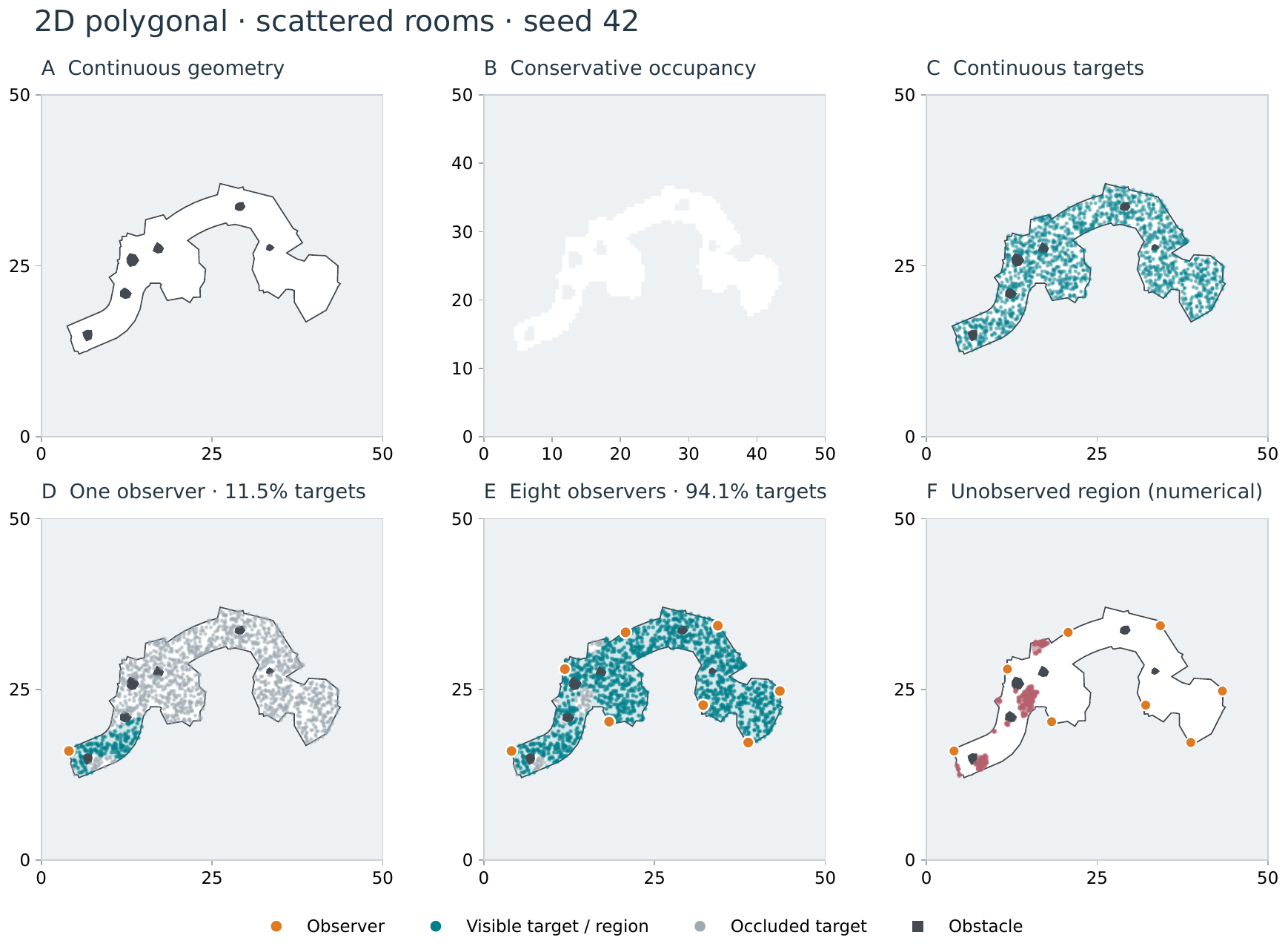}{2D polygonal, $50\times50$ world units, scattered rooms, seed 42: 10 realized rooms, 12 original layout paths, 6 inserted obstacles, graph cycle rank 3, and 6 polygon holes. Occupancy uses $100\times100$ cells. Free area 339.6 square units. The 1,500 interior targets and eight observers give 94.1\% union target coverage using polygon-edge visibility. }

Two more seeds change the family and realized world.

\atlasfigure[.33\textheight]{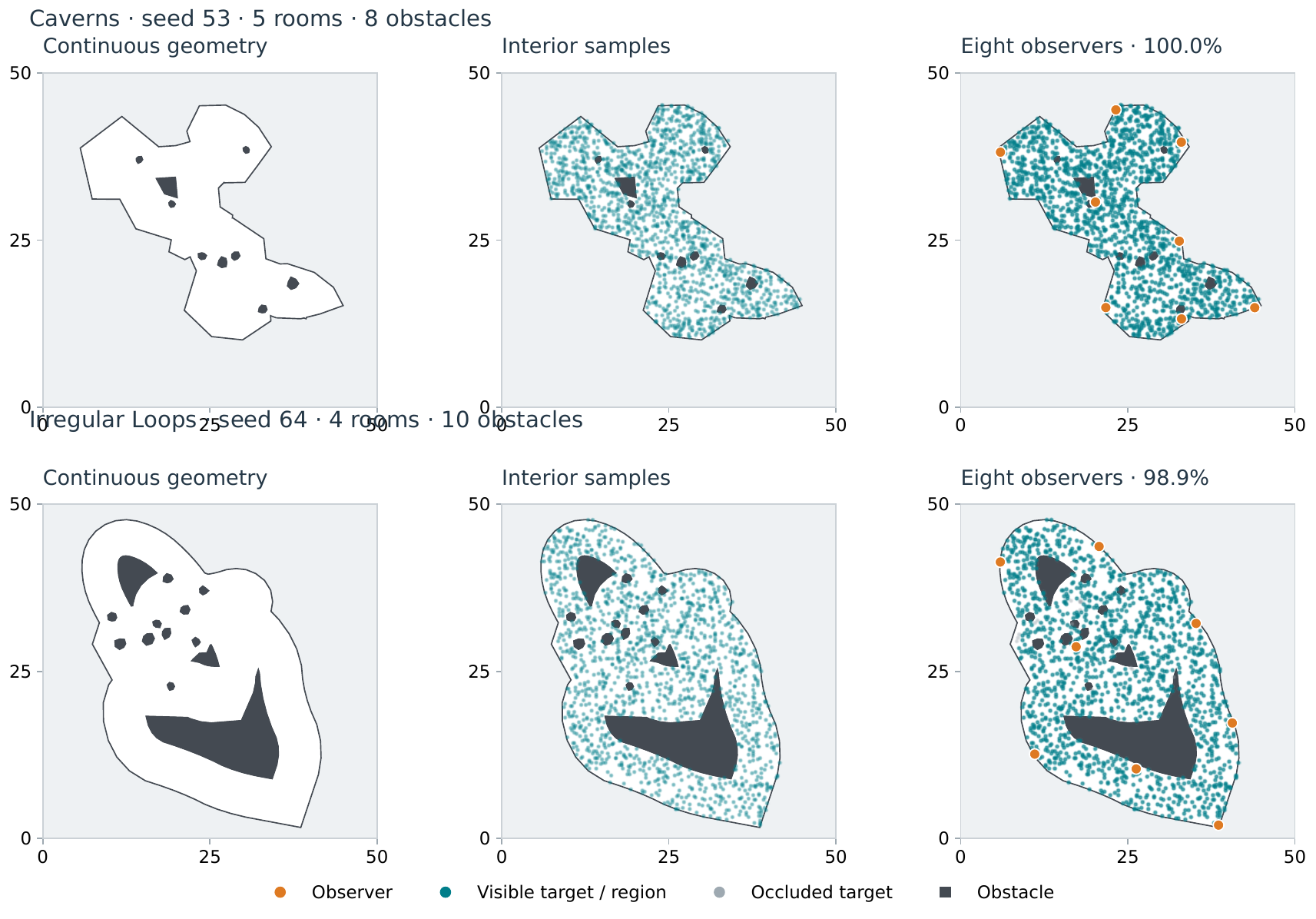}{2D polygonal, $50\times50$ world units, caverns, seed 53: 5 realized rooms, 6 original layout paths, 8 inserted obstacles, graph cycle rank 2, and 9 polygon holes. Occupancy uses $100\times100$ cells. Free area 589.8 square units. The 1,500 interior targets and eight observers give 100.0\% union target coverage using polygon-edge visibility.  2D polygonal, $50\times50$ world units, irregular loops, seed 64: 4 realized rooms, 5 original layout paths, 10 inserted obstacles, graph cycle rank 3, and 13 polygon holes. Occupancy uses $100\times100$ cells. Free area 950.8 square units. The 1,500 interior targets and eight observers give 98.9\% union target coverage using polygon-edge visibility. }

\FloatBarrier
\subsection{Curvilinear worlds in two dimensions}

\atlasfigure[.33\textheight]{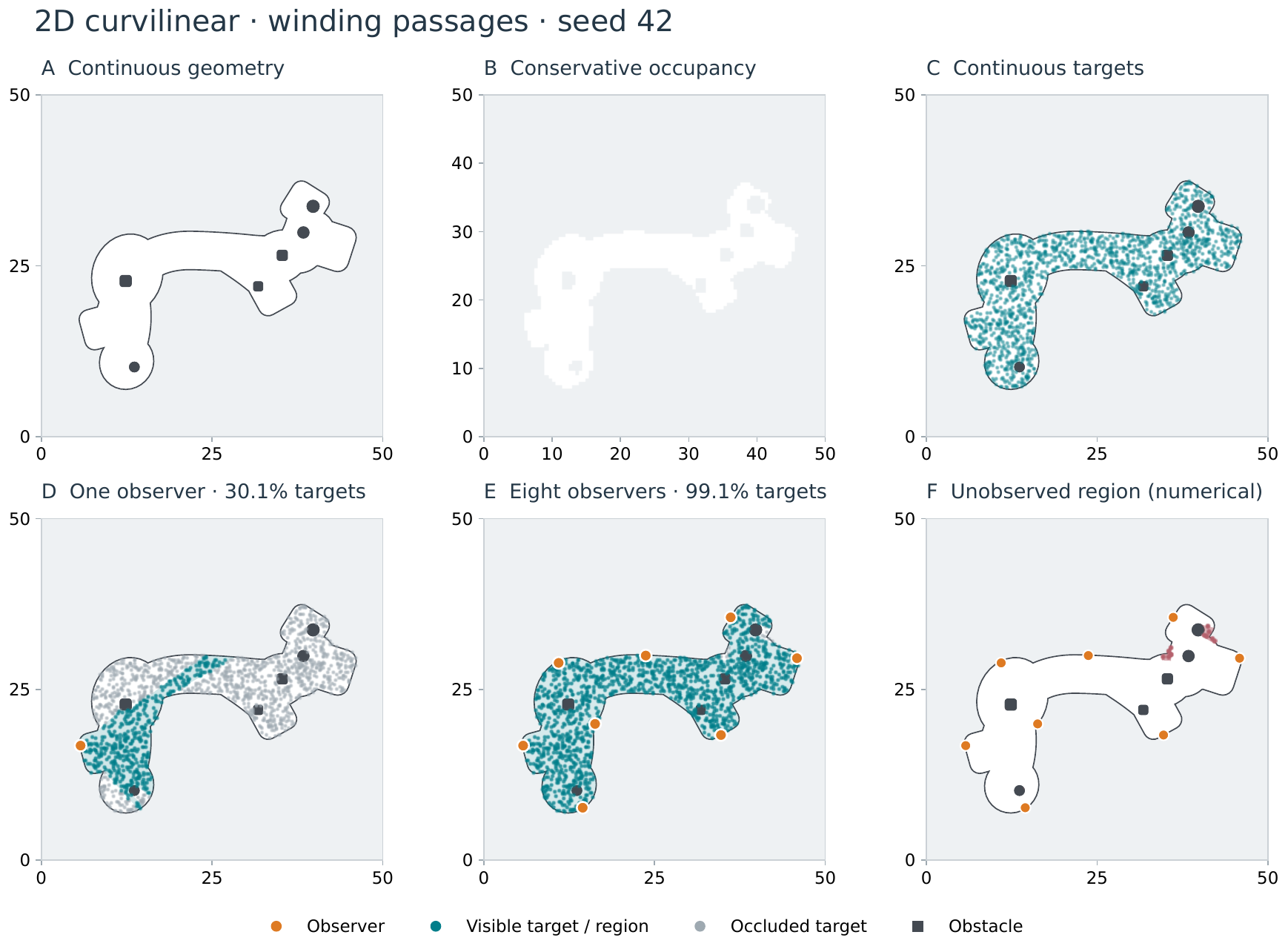}{2D curvilinear, $50\times50$ world units, winding passages, seed 42: 8 realized rooms, 5 original layout paths, 6 inserted obstacles, graph cycle rank 0, and 6 polygon holes. Occupancy uses $100\times100$ cells. Free area 427.7 square units. The 1,500 interior targets and eight observers give 99.1\% union target coverage using polygon-edge visibility. }

Two more seeds change the family and realized world.

\atlasfigure[.33\textheight]{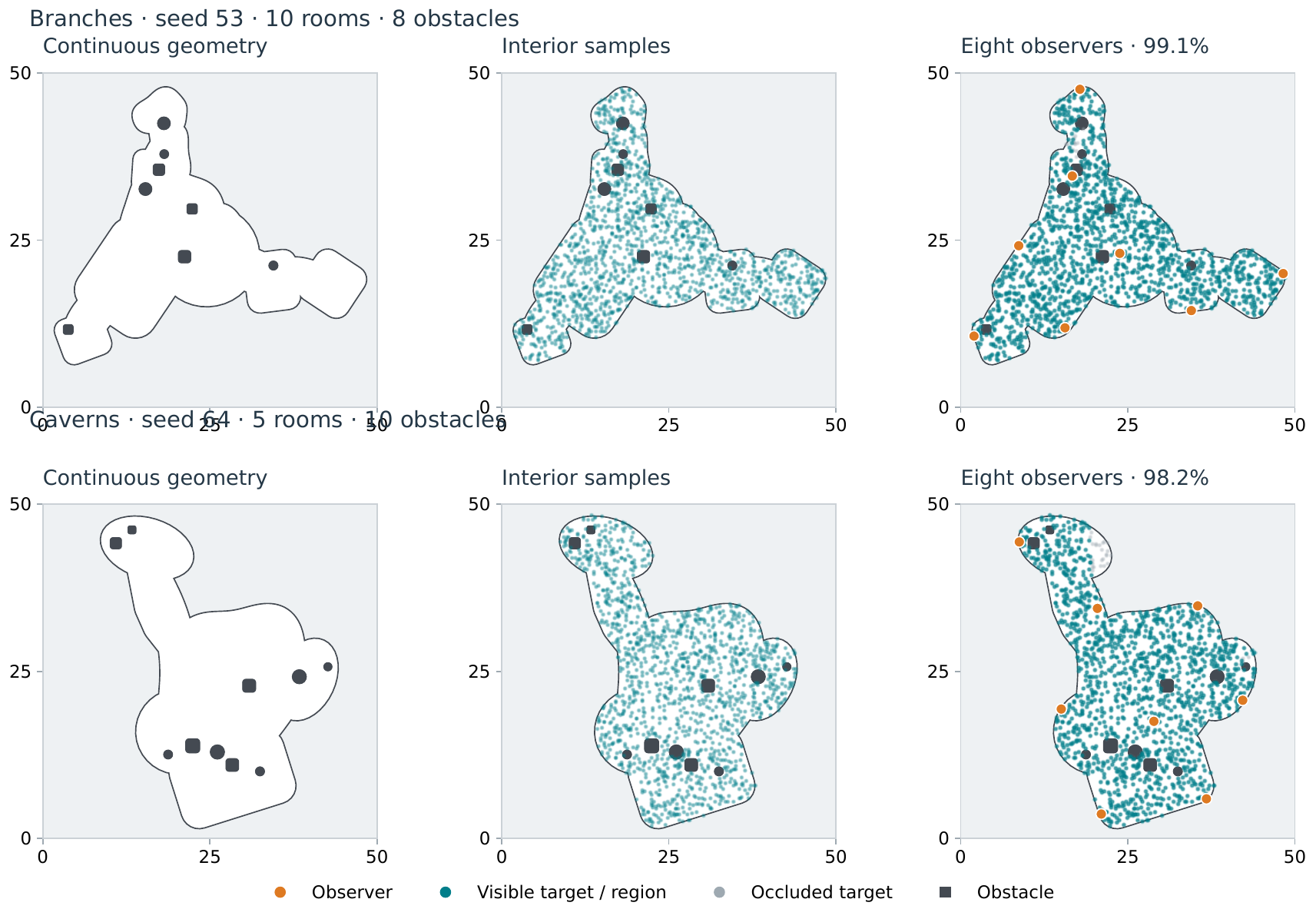}{2D curvilinear, $50\times50$ world units, branches, seed 53: 10 realized rooms, 9 original layout paths, 8 inserted obstacles, graph cycle rank 0, and 8 polygon holes. Occupancy uses $100\times100$ cells. Free area 704.4 square units. The 1,500 interior targets and eight observers give 99.1\% union target coverage using polygon-edge visibility.  2D curvilinear, $50\times50$ world units, caverns, seed 64: 5 realized rooms, 7 original layout paths, 10 inserted obstacles, graph cycle rank 3, and 10 polygon holes. Occupancy uses $100\times100$ cells. Free area 824.7 square units. The 1,500 interior targets and eight observers give 98.2\% union target coverage using polygon-edge visibility. }

\FloatBarrier
\subsection{Orthogonal worlds in three dimensions}

\atlasfigure[.33\textheight]{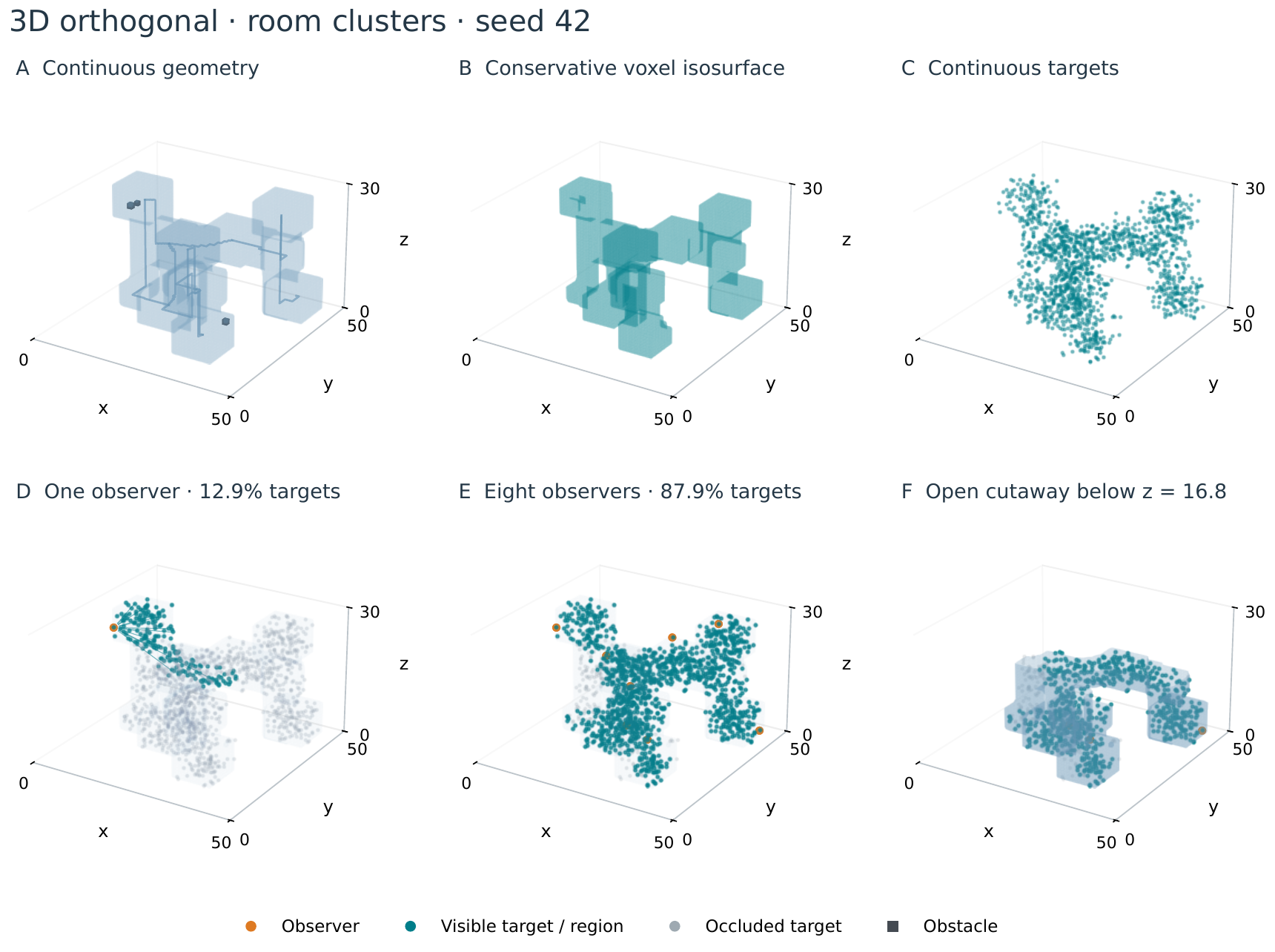}{3D orthogonal, $50\times50\times30$ world units, room clusters, seed 42: 9 realized rooms, 10 original layout paths, 3 inserted obstacles, graph cycle rank 2, and 3 protected internal cavities from inserted obstacles. Occupancy uses $70\times70\times42$ voxels. Estimated free volume 10123$\pm$105 cubic units (one standard error). The 1,500 interior targets and eight observers give 87.9\% union target coverage using analytic constructive-solid visibility. }

Two more seeds change the family and realized world.

\atlasfigure[.33\textheight]{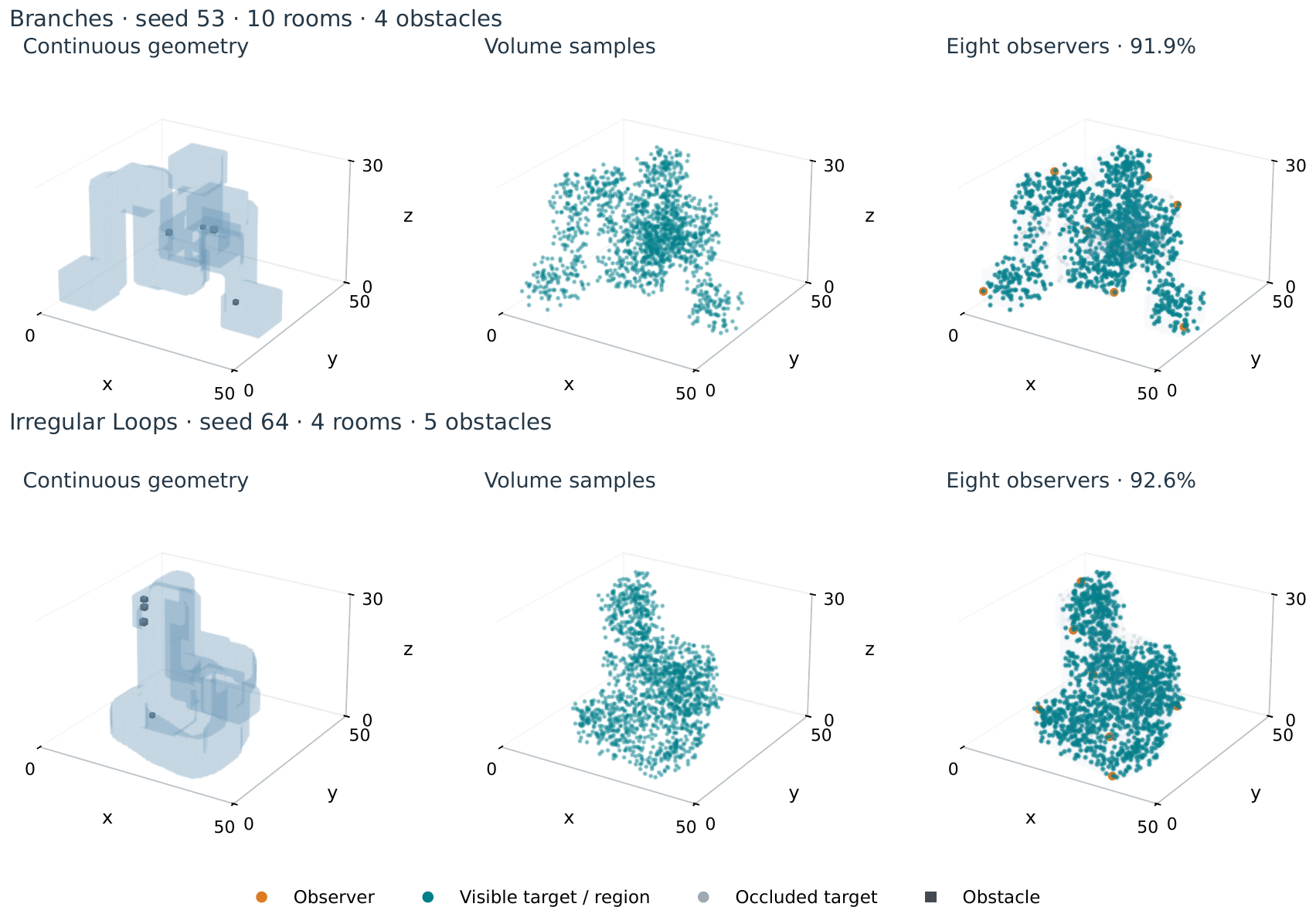}{3D orthogonal, $50\times50\times30$ world units, branches, seed 53: 10 realized rooms, 9 original layout paths, 4 inserted obstacles, graph cycle rank 0, and 4 protected internal cavities from inserted obstacles. Occupancy uses $70\times70\times42$ voxels. Estimated free volume 11644$\pm$111 cubic units (one standard error). The 1,500 interior targets and eight observers give 91.9\% union target coverage using analytic constructive-solid visibility.  3D orthogonal, $50\times50\times30$ world units, irregular loops, seed 64: 4 realized rooms, 5 original layout paths, 5 inserted obstacles, graph cycle rank 3, and 5 protected internal cavities from inserted obstacles. Occupancy uses $70\times70\times42$ voxels. Estimated free volume 11362$\pm$110 cubic units (one standard error). The 1,500 interior targets and eight observers give 92.6\% union target coverage using analytic constructive-solid visibility. }

\FloatBarrier
\subsection{Polyhedral worlds in three dimensions}

\atlasfigure[.33\textheight]{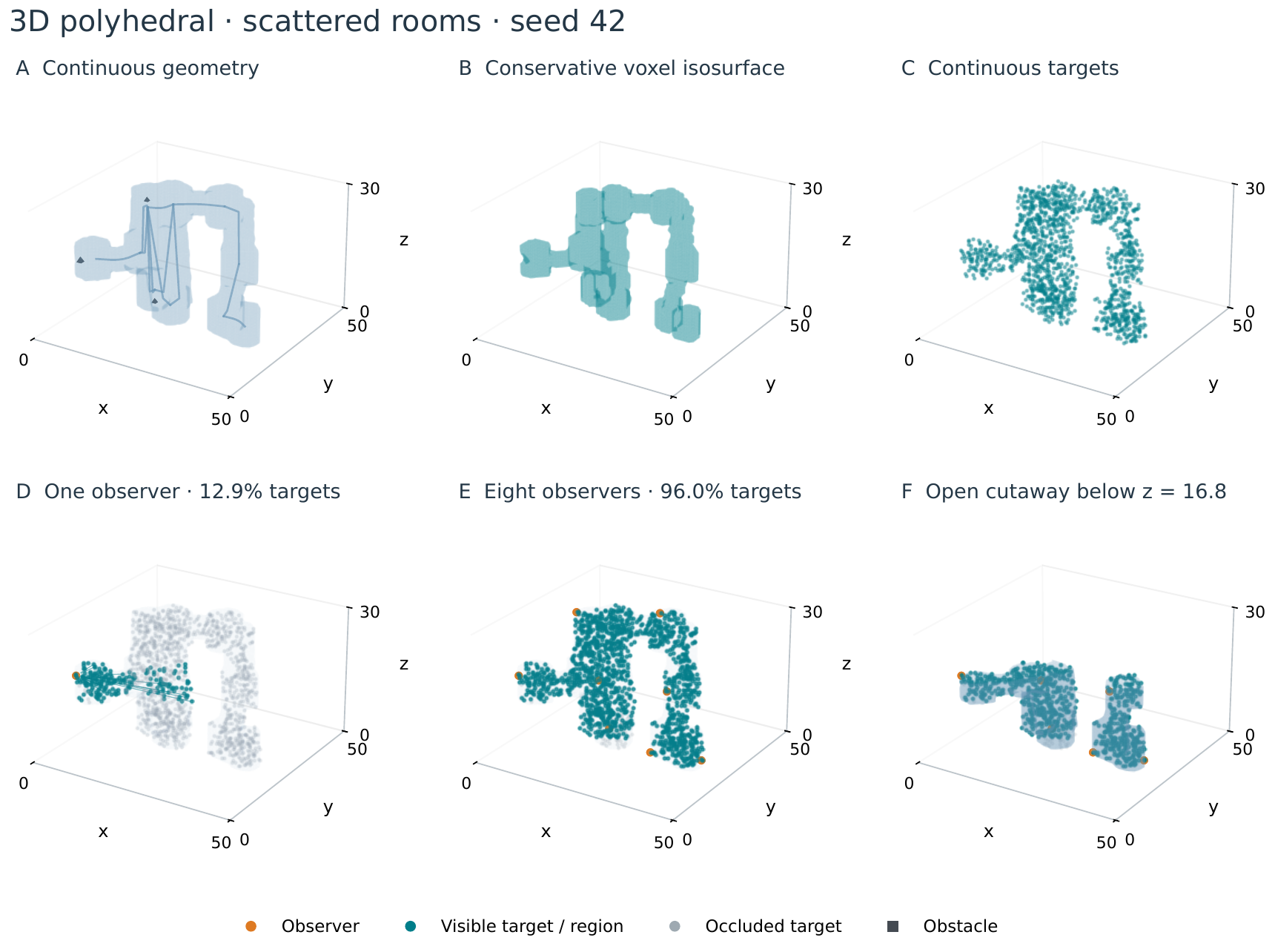}{3D polyhedral, $50\times50\times30$ world units, scattered rooms, seed 42: 10 realized rooms, 12 original layout paths, 3 inserted obstacles, graph cycle rank 3, and 3 protected internal cavities from inserted obstacles. Occupancy uses $70\times70\times42$ voxels. Estimated free volume 4852$\pm$75 cubic units (one standard error). The 1,500 interior targets and eight observers give 96.0\% union target coverage using analytic constructive-solid visibility. }

Two more seeds change the family and realized world.

\atlasfigure[.33\textheight]{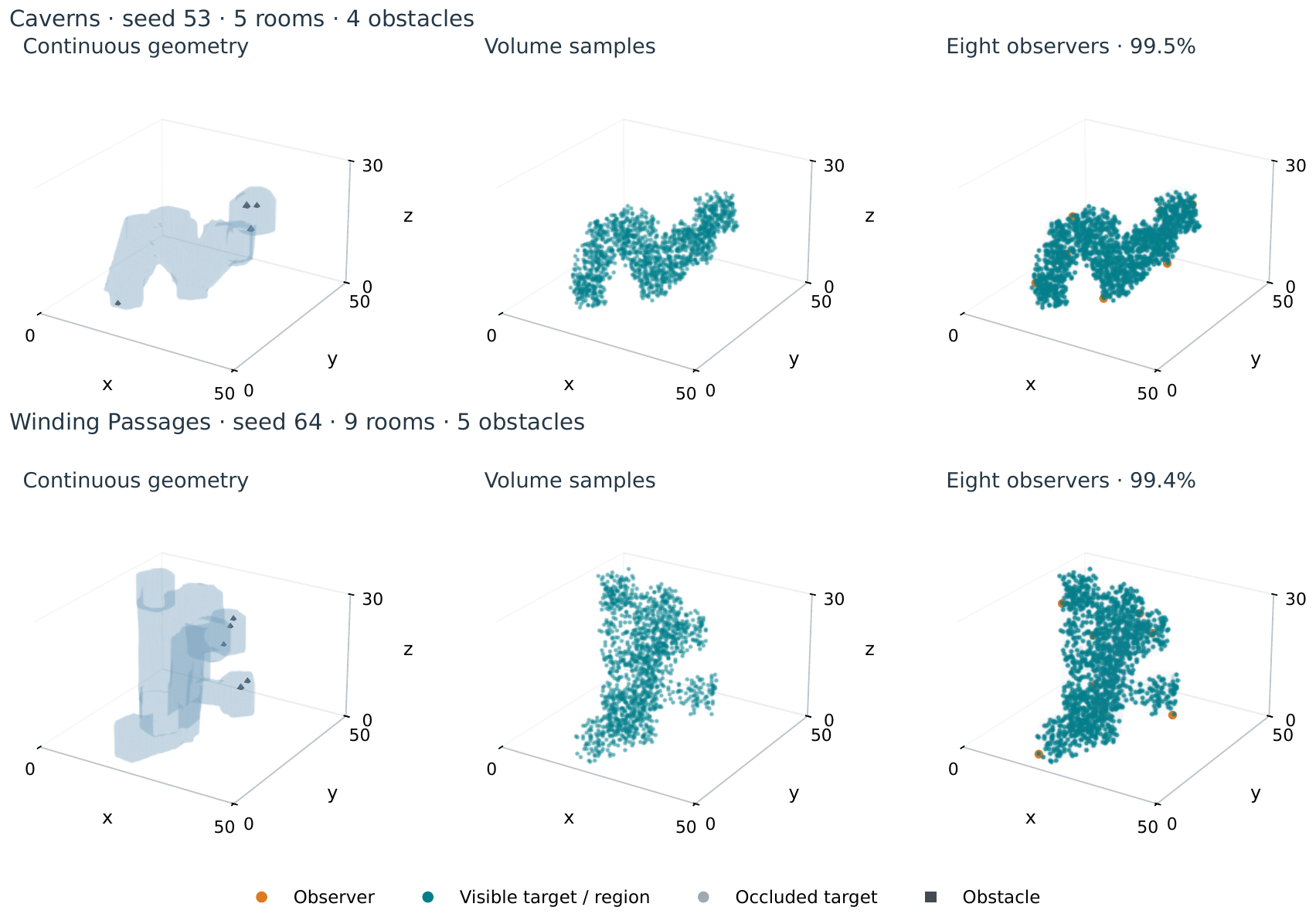}{3D polyhedral, $50\times50\times30$ world units, caverns, seed 53: 5 realized rooms, 6 original layout paths, 4 inserted obstacles, graph cycle rank 2, and 4 protected internal cavities from inserted obstacles. Occupancy uses $70\times70\times42$ voxels. Estimated free volume 3542$\pm$65 cubic units (one standard error). The 1,500 interior targets and eight observers give 99.5\% union target coverage using analytic constructive-solid visibility.  3D polyhedral, $50\times50\times30$ world units, winding passages, seed 64: 9 realized rooms, 6 original layout paths, 5 inserted obstacles, graph cycle rank 0, and 5 protected internal cavities from inserted obstacles. Occupancy uses $70\times70\times42$ voxels. Estimated free volume 7342$\pm$91 cubic units (one standard error). The 1,500 interior targets and eight observers give 99.4\% union target coverage using analytic constructive-solid visibility. }

\FloatBarrier
\subsection{Curvilinear worlds in three dimensions}

\atlasfigure[.33\textheight]{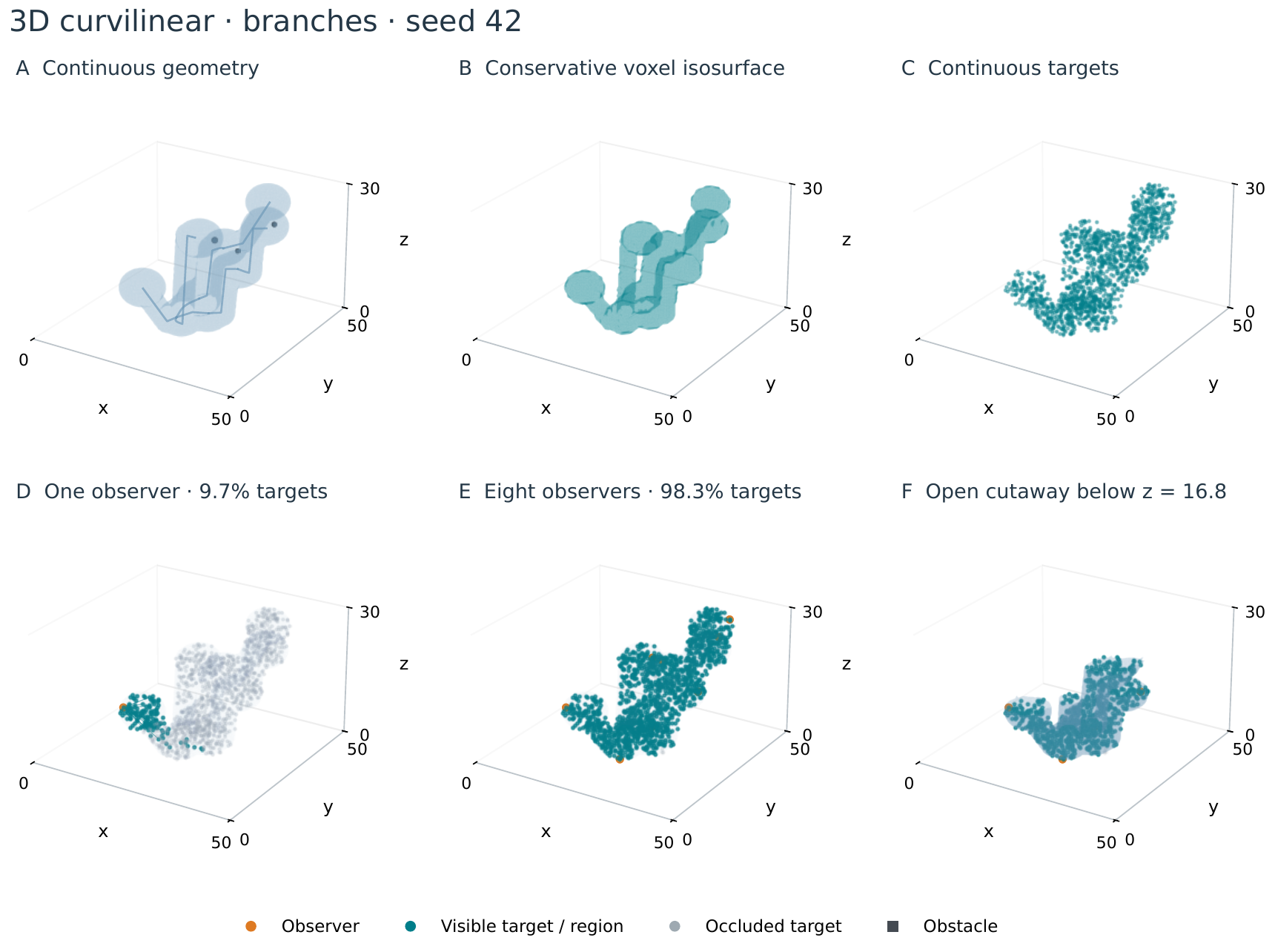}{3D curvilinear, $50\times50\times30$ world units, branches, seed 42: 9 realized rooms, 8 original layout paths, 3 inserted obstacles, graph cycle rank 0, and 3 protected internal cavities from inserted obstacles. Occupancy uses $70\times70\times42$ voxels. Estimated free volume 4820$\pm$75 cubic units (one standard error). The 1,500 interior targets and eight observers give 98.3\% union target coverage using analytic constructive-solid visibility. }

Two more seeds change the family and realized world.

\atlasfigure[.33\textheight]{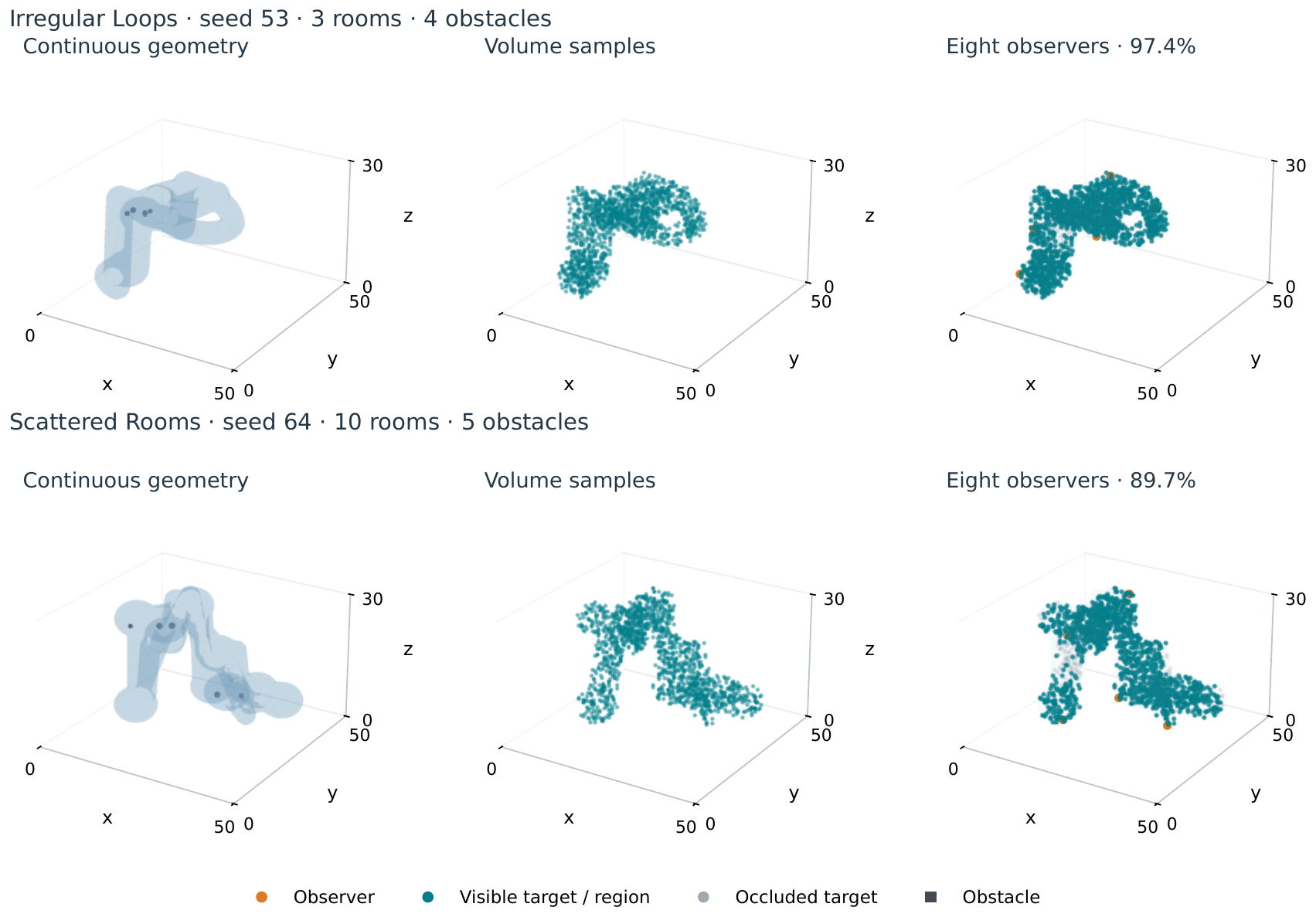}{3D curvilinear, $50\times50\times30$ world units, irregular loops, seed 53: 3 realized rooms, 3 original layout paths, 4 inserted obstacles, graph cycle rank 2, and 4 protected internal cavities from inserted obstacles. Occupancy uses $70\times70\times42$ voxels. Estimated free volume 4385$\pm$72 cubic units (one standard error). The 1,500 interior targets and eight observers give 97.4\% union target coverage using analytic constructive-solid visibility.  3D curvilinear, $50\times50\times30$ world units, scattered rooms, seed 64: 10 realized rooms, 17 original layout paths, 5 inserted obstacles, graph cycle rank 8, and 5 protected internal cavities from inserted obstacles. Occupancy uses $70\times70\times42$ voxels. Estimated free volume 6599$\pm$87 cubic units (one standard error). The 1,500 interior targets and eight observers give 89.7\% union target coverage using analytic constructive-solid visibility. }



\section{Controlled geometry and visibility comparisons}
\label{sec:comparisons}
These earlier experiments change recorded inputs and retain the resulting
geometry and visibility. Continuous target sets are stated in each caption.
The archived grid-visibility comparison in Section~7.3 uses common
continuous coordinates under its original query convention. It is retained
as evidence about wall representation; new discrete queries use free-cell
or free-voxel centers as defined in Section~4.6.

\FloatBarrier
\subsection{Layout realization and obstacle insertion}
A shared room-and-passage layout permits changes to wall geometry while centers, original passage paths and the construction graph remain fixed. The resulting free space and visibility can change.

\atlasfigure[.32\textheight]{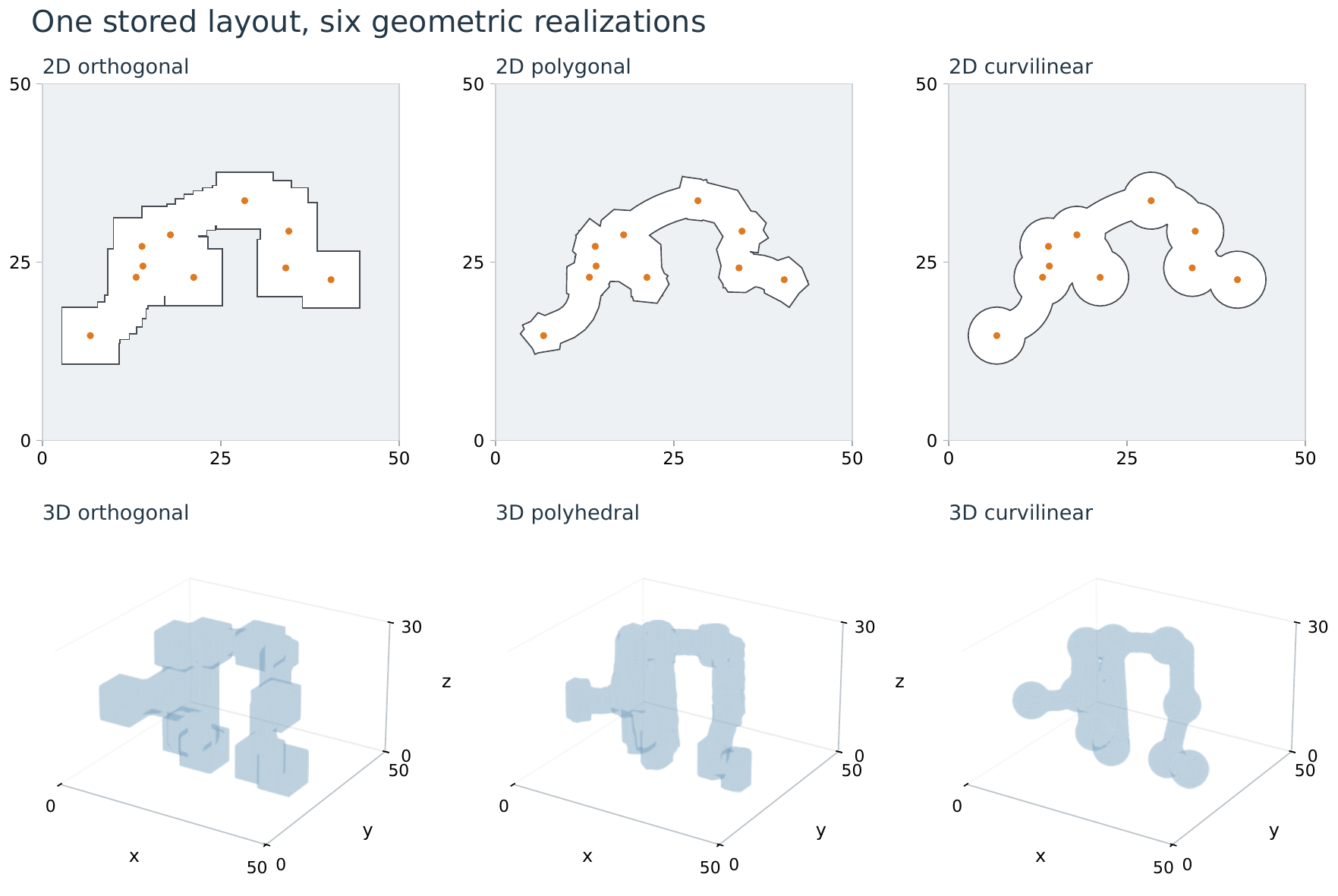}{One layout realized in all six classes. The three 2D and three multilevel 3D constructions share an identical saved layout artifact. Axis alignment, oblique planar walls, and curved chamber surfaces change independently of graph identity. Areas, volumes and sensing behavior need not match. These examples contain no inserted obstacles.}

\atlasfigure[.32\textheight]{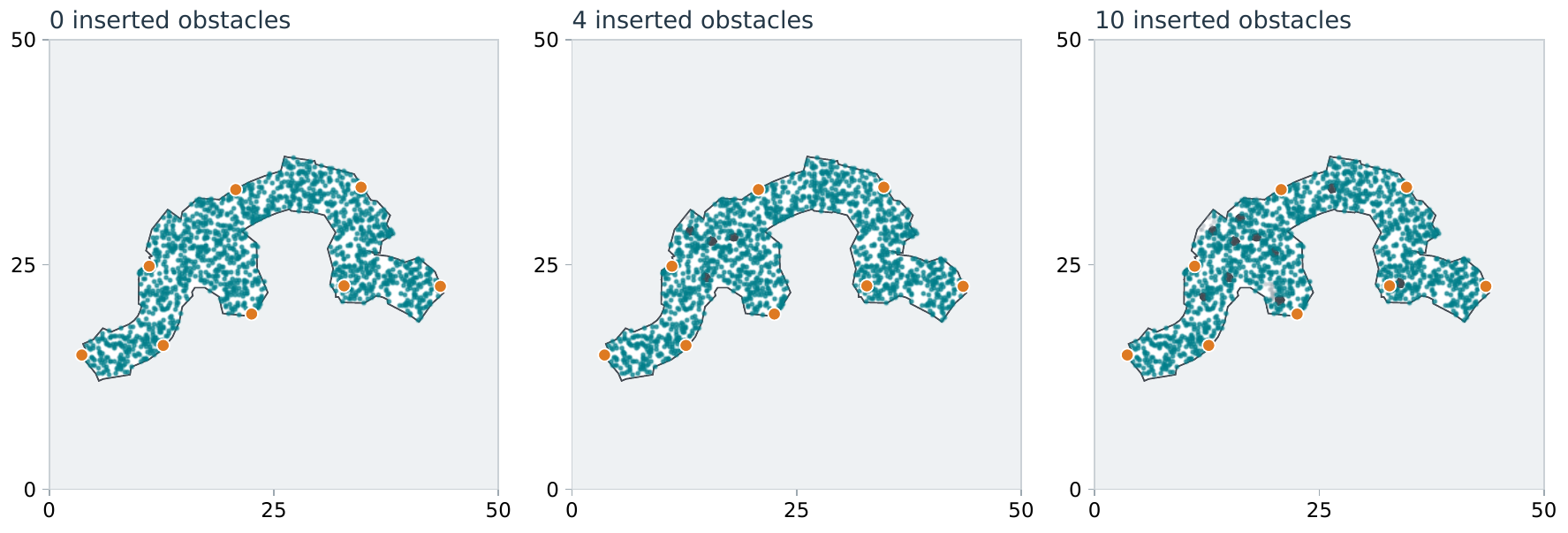}{Zero, four and ten irregular polygon obstacles with an unchanged seed, stored layout, room realization and passage realization. A $200^2$ conservative grid and fixed realization attempt keep this study controlled. The free-space polygons are checked to be nested as obstacles are added. Target sets are resampled in each resulting free space; this panel is not a matched-target coverage comparison.}

\FloatBarrier
\subsection{Observer position and accumulated coverage}
Observer position changes the visible target set in a fixed world. Nested prefixes of one saved sequence isolate the effect of adding observers; these viewpoints are illustrative.

\atlasfigure[.32\textheight]{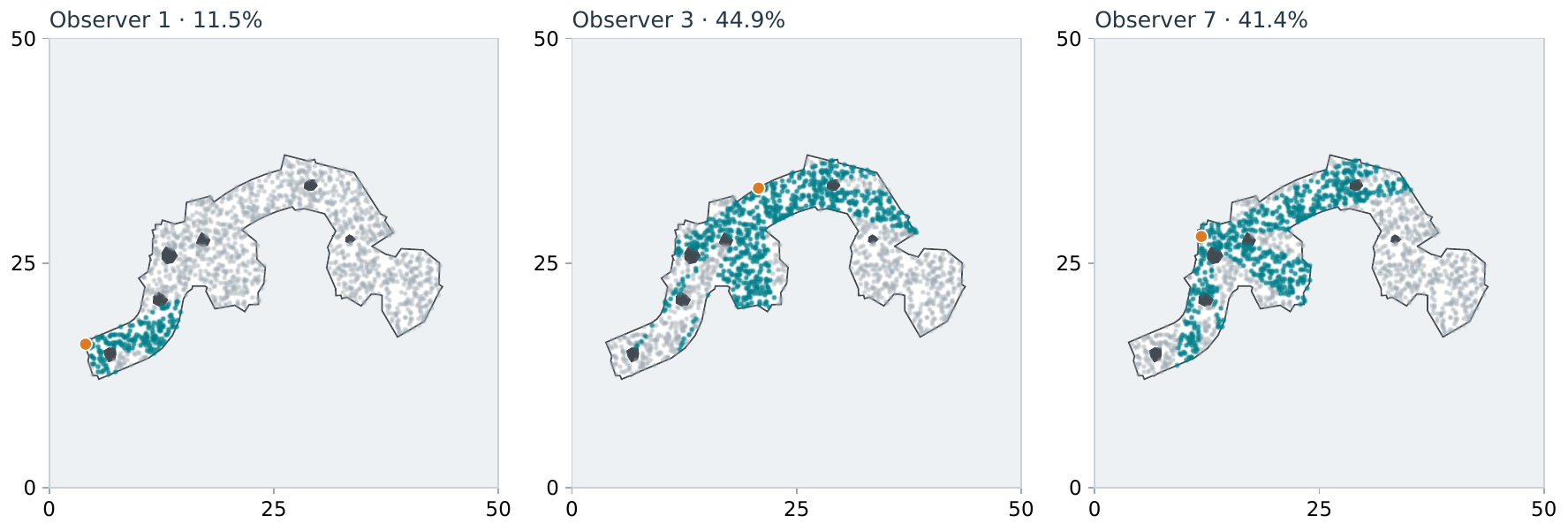}{Three of the saved observer positions in one polygonal world. Colors come from independent Rust rows against the same 1,500 targets. Each panel reports one observer's target fraction. The chosen positions are illustrative and do not optimize coverage.}

\atlasfigure[.32\textheight]{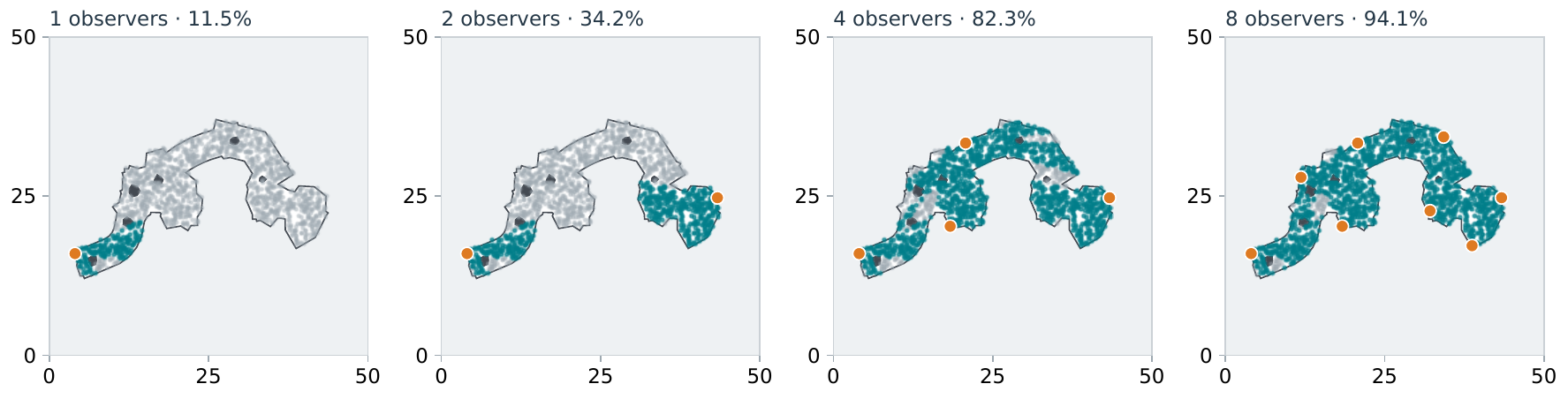}{One, two, four and eight observers from the same deterministic sequence. The visible union is computed from the corresponding CSR rows. Coverage is monotonic for these nested sets but is a finite-target quantity; the sequence does not establish an optimal observer count.}

\FloatBarrier
\subsection{Representation effects on visibility}
The same straight sight segment is checked against different wall representations. Conservative cells can block a segment that remains inside the continuous world. Three-dimensional queries also retain elevation and occlusion.

\atlasfigure[.32\textheight]{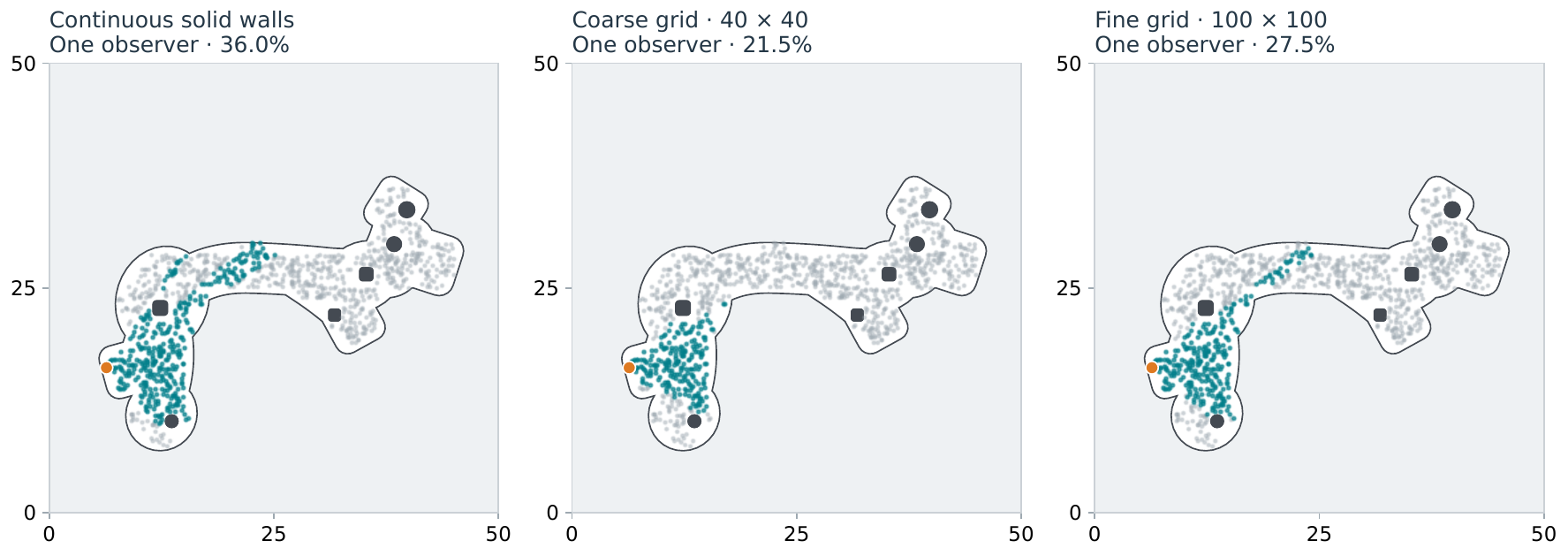}{The same observers and 1046 targets, all lying in continuous free space and inside free cells of both grids. These archived targets retain their continuous coordinates; they are not snapped to cell centers. One observer sees 36.0\% continuously, 21.5\% with coarse cells and 27.5\% with fine cells. All classifications are computed, and disagreement is retained rather than forced away.}

\atlasfigure[.32\textheight]{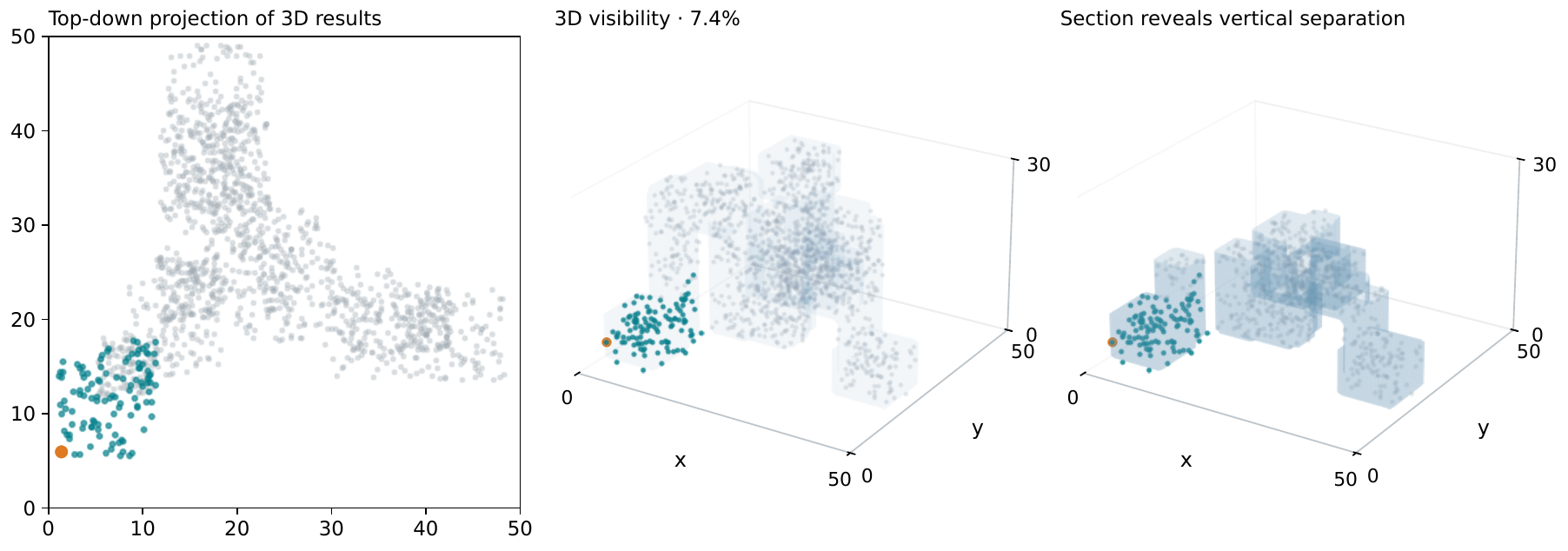}{A multilevel orthogonal world, with top-down projection, full 3D target visibility, and a cutaway. The top-down colors are projections of the actual 3D results, not an independent planar visibility calculation. Targets close in projected coordinates can occupy different elevations with intervening empty or blocked volume. The observer and all displayed target classifications are saved.}

\FloatBarrier
\subsection{Physical scale and experimental units}
Physical extent, cell pitch and array dimensions are different controls. These examples scale saved geometry together with observer and target coordinates, preserving relative structure rather than adding rooms or passages.

\atlasfigure[.32\textheight]{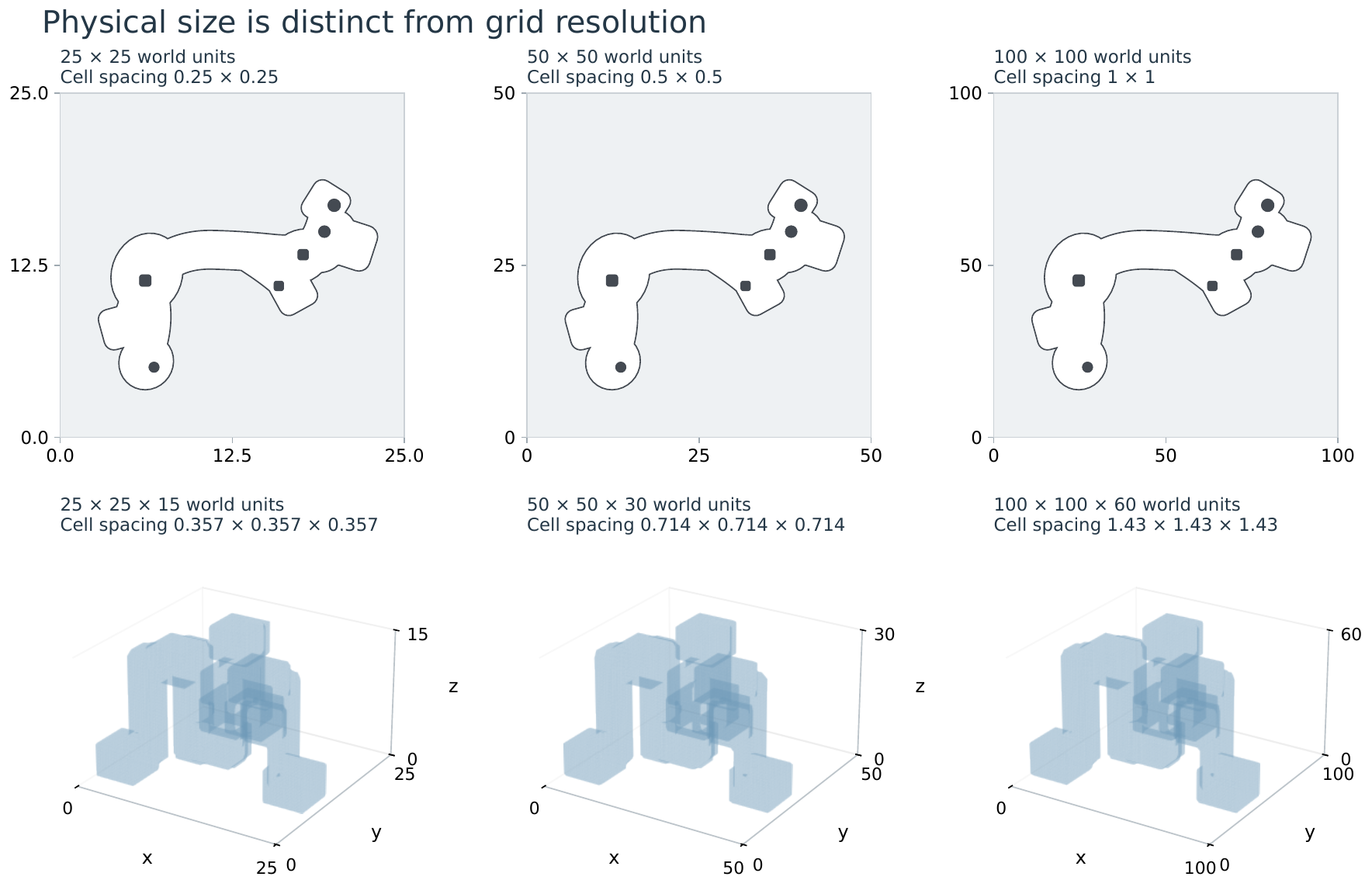}{Domains of $25\times25$, $50\times50$, and $100\times100$ units in 2D; $25\times25\times15$, $50\times50\times30$, and $100\times100\times60$ units in 3D. Each row retains its array shape and normalized structure; the physical cell spacing changes. All six visibility computations use the correspondingly scaled eight observers and 1,500 targets. The computed matrices have zero disagreements with their source matrices under unlimited-range sensing.}

\begin{center}\small\begin{tabular}{@{}lrrrr@{}}\toprule & Domain extent & Array shape & Cell pitch & Free measure\\\midrule

2D & $25\times25$ & $100\times100$ & 0.250 & 106.9 \\
2D & $50\times50$ & $100\times100$ & 0.500 & 427.7 \\
2D & $100\times100$ & $100\times100$ & 1.000 & 1710.7 \\
3D & $25\times25\times15$ & $70\times70\times42$ & 0.357 & 1455$\pm$14 \\
3D & $50\times50\times30$ & $70\times70\times42$ & 0.714 & 11644$\pm$111 \\
3D & $100\times100\times60$ & $70\times70\times42$ & 1.429 & 93150$\pm$887 \\

\bottomrule\end{tabular}\end{center}

The free measure is area in 2D and estimated volume with one standard error in 3D; the source Monte Carlo volume and its standard error scale by the cube of the similarity factor. Cell pitch is isotropic in these examples. Doubling extent multiplies area by four and volume by eight, but does not add rooms, obstacles or graph vertices. Keeping the planar pitch at $0.5$ units instead requires a $200\times200$ array for the $100\times100$ domain; that derived mask is saved separately. A corresponding fixed-pitch doubling in 3D would multiply the voxel count by eight. Finite sensor ranges require separate attention to physical scale.



\section{Software components and persistent experiment inputs}
\label{sec:software}
The software provides world generation and a visibility backend for
vision-based online tasks. Geometry generation, occupancy conversion,
sampling and sensing have separate inputs and outputs. A saved visibility query can be replayed
without rebuilding a world; a new resolution can be derived from unchanged
continuous geometry. The research bundle retains code, scientific inputs and provenance; the
Overleaf bundle retains the manuscript and required figures.

\begin{center}\small
\begin{tabularx}{\linewidth}{@{}l>{\raggedright\arraybackslash}X@{}}
\toprule
Component & Responsibility and persistent output\\
\midrule
\code{layout.py} & Bundled layout families; room templates, passage paths
and the construction graph in \code{layout.json}.\\
\code{layout\_distributions.py} & Configurable family controls and joint room-placement
laws; individual layout draws and fitting transforms.\\
\code{parameters.py} & Typed distribution specifications, sampled parameters
and recorded retry streams.\\
\code{generation.py} & Boundary realization, obstacle insertion, occupancy,
GeoJSON or solids and generation metadata.\\
\code{solid.py} & 3D membership and clearance, primitive separation, overlap witnesses
and clipping references.\\
\code{sampling.py} & Continuous samples and their seeds; all free grid centers for discrete
queries, adjusted observers and volume estimates.\\
\code{visibility.py} & Exact Rust query inputs, compact CSR results,
sparse matrices and target unions.\\
\code{hierarchy.py} & Exact mask-tree counts, depth profiles and leaf bounds;
\code{hierarchy\_study.py} derives the refinement study.\\
Validation and figures & Independent numerical checks, saved-world audits,
display exports and measurement-derived captions.\\
\bottomrule
\end{tabularx}
\end{center}

\subsection{Replayable sensing contract}
A Rust query records the dimension, mode, ordered observers and targets,
numerical tolerance and sensor settings. It includes polygon rings for
continuous 2D queries, positive and negative primitives for 3D queries, or
the array shape, origin, spacings and flattened mask for discrete queries.
The standalone executable accepts \code{--input}, \code{--output} and
\code{--threads}; it replays these coordinates and settings independently
of world generation. Every showcase retains its continuous input JSON.
An online algorithm can submit newly selected observer positions against
the same world and target convention, and retain its accumulated observations.

\subsection{Visible and uncovered target sets}
The compact CSR starts with the eight bytes \code{WGVCSR01}, followed by
little-endian unsigned 64-bit row, column and nonzero counts:
\(M\) observers, \(N\) targets and \(K\) visible pairs.
Then \(M+1\) offsets delimit \(K\) target indices.
Every stored pair is true, so the file occupies
\[
 32+8(M+1+K)\ \text{bytes}.
\]
This layout stores explicit visible sets. It can cost more bytes than a
small dense Boolean matrix when many targets are visible; sparsity and
bytes per index both matter.

Python validates the CSR structure and saves a sparse \code{.npz}, row
counts, the visible union and its unobserved complement. Coordinate arrays
identify every row and column. Timing reports separate the Rust query loop
from subprocess and conversion time.


\section{Discussion and conclusion}
\label{sec:conclusion}
For vision-based online tasks, the framework separates the room-and-passage
layout, continuous free
space, conservative occupancy, observation targets and saved visibility
results. Six construction families can be realized as orthogonal,
polygonal or polyhedral, and curvilinear worlds in two or three dimensions.
Fixed values, uniform ranges and supported probability laws control
the exposed generation parameters. Requested settings, realized draws,
fitting transforms and accepted attempts remain inspectable in the
saved artifacts.

Continuous geometry and discrete occupancy describe the same generated
world at different levels of resolution. Whole-cell or whole-voxel
certificates protect the free grid from admitting wall-crossing cells.
Continuous sensing uses sampled target coordinates; current discrete
sensing uses every free cell or voxel center. Saved query inputs identify
the wall representation, target positions, observers and numerical
conventions needed to replay a visibility result.

The timing experiment in Section~\ref{sec:generation-reproducibility}
measures complete generation calls for six configurations. Five timed
repeats and an additional fresh-process replay per configuration reproduce
the geometry, grids and construction records exactly in the recorded
environment. Generation times vary even when the resulting world is
identical. Source hashes, configurations and dependency versions are
therefore retained alongside measured times.

The restored atlas and comparisons show how wall geometry, obstacles,
observer placement, grid resolution and physical scale affect the
representations and measured visibility. These examples preserve their
recorded target sets. Target fractions from different sampling conventions
or different worlds require their denominators and coordinates to be
considered explicitly.

Appendix~\ref{app:hierarchy} explains the quadtree and octree measure of
world complexity, then uses POCGAGP as an example with illustrative worlds
and prior experimental evidence. The generated environments and replayable
sensing inputs support online algorithms that choose observations and
retain the information accumulated during an experiment.


\FloatBarrier


\FloatBarrier
\appendix

\section{Complexity of worlds}
\label{sec:complexity-main}
\label{sec:complexity}
\label{app:hierarchy}

\newcommand{\hierarchyScalingTable}{\begin{center}\small\begin{tabular}{@{}llrrrrr@{}}\toprule
Example & \(n\) range & \(T_1\) & \(T_2\) & \(T_3\) & \(T_4\) & \(T_4/T_3\)\\\midrule
2D orthogonal & 32--256 & 217 & 717 & 1,365 & 2,389 & 1.75 \\
2D polygonal & 32--256 & 201 & 581 & 1,349 & 2,837 & 2.10 \\
2D curvilinear & 32--256 & 233 & 569 & 1,317 & 2,693 & 2.04 \\
3D orthogonal & 16--128 & 209 & 2,337 & 11,577 & 53,057 & 4.58 \\
3D polyhedral & 16--128 & 73 & 1,465 & 7,897 & 33,145 & 4.20 \\
3D curvilinear & 16--128 & 113 & 1,465 & 7,993 & 38,153 & 4.77 \\\bottomrule\end{tabular}\end{center}}
\newcommand{\hierarchyAnalyticTable}{\begin{center}\small\begin{tabular}{@{}rrrrr@{}}\toprule
\(n\) & 2D slab & 2D alternating & 3D slab & 3D alternating\\\midrule
8 & 5 & 85 & 9 & 585\\
16 & 5 & 341 & 9 & 4,681\\
32 & 5 & 1,365 & 9 & 37,449\\
64 & 5 & 5,461 & 9 & 299,593\\\bottomrule\end{tabular}\end{center}}
\newcommand{\hierarchyLegacyTable}{\begin{center}\small\begin{tabular}{@{}rrrr@{}}\toprule
Source grid side & Stored legacy \(T\) & Recomputed \(T\) & Difference\\\midrule
50 & 17,825 & 17,825 & +0 \\
100 & 43,757 & 43,757 & +0 \\
250 & 72,577 & 72,577 & +0 \\\bottomrule\end{tabular}\end{center}}
\newcommand{\hierarchyNativeTable}{\begin{center}\small\begin{tabular}{@{}lrrr@{}}\toprule
Class & Example 1 & Example 2 & Example 3\\\midrule
2D orthogonal & 1,033 & 1,265 & 1,373\\
2D polygonal & 1,049 & 1,141 & 1,693\\
2D curvilinear & 1,065 & 1,329 & 1,229\\
3D orthogonal & 13,105 & 16,041 & 17,289\\
3D polyhedral & 8,113 & 6,537 & 11,457\\
3D curvilinear & 9,905 & 9,433 & 13,841\\\bottomrule\end{tabular}\end{center}}
\newcommand{\hierarchyFineRatios}{The finest doubling multiplies 2D node counts by 1.75--2.10, versus four times the dense entries. In 3D the corresponding node-count factors are 4.20--4.77, versus eight times the dense entries.}
\newcommand{\hierarchyRotationCounts}{805, 597, 709, 725}
\newcommand{\hierarchyObstacleCounts}{1,725, 1,869, 2,081}


World complexity depends on what is being measured. A larger open room
can require more travel or more sensing positions under a finite sensor
range. More corners can increase the number of boundary elements processed
by a geometric algorithm. These descriptors can therefore be relevant in
some settings, but neither determines difficulty for every task.

We use the size of a region quadtree in 2D or a region octree in 3D as a
measure of spatial detail at a declared resolution~\cite{samet,meagher}.
The hierarchy records how much subdivision is needed to represent free
and blocked space. It gives a reproducible ranking under a common domain,
pitch, root and orientation convention. Its relationship to task outcomes
is evaluated separately: Section~\ref{app:pocgagp} uses POCGAGP as an example,
with Figure~\ref{fig:complexity-counterexamples} illustrating why physical
size and corner count alone are insufficient for its visibility component.

\subsection{Constructing the hierarchy}
\label{sec:complexity-construction}
Start with one square containing the 2D occupancy mask, or one cube
containing the 3D mask. If all entries in a region are free, that region
becomes a free leaf. If all are blocked, it becomes a blocked leaf.
Otherwise, split the region into four equal squares in 2D or eight equal
cubes in 3D, and repeat. Subdivision stops when each retained region is
homogeneous; an individual finest-grid cell or voxel is always homogeneous.
This hierarchy preserves the saved Boolean mask exactly.

Large uniform regions need few nodes. Boundaries, small obstacles and
narrow passages can require finer subdivision and therefore more nodes.
The classification uses the conservative occupancy defined in Section~2:
a blocked entry can include a cell that is partly free in continuous space.
The hierarchy ranks this discrete representation of the world.

\subsection{The score and ranking rule}
\label{sec:complexity-ranking}
Let \(I\) be the number of subdivided internal nodes and \(L\) the number
of leaves. Count every retained node, including free leaves, blocked
leaves and internal nodes. The complexity score is
\[
 C_h(\W)=T=I+L=
 \begin{cases}
  1+4I, & \text{quadtree in 2D},\\
  1+8I, & \text{octree in 3D},
 \end{cases}
\]
where \(h\) is the finest cell or voxel pitch. These identities hold because
each subdivision retains all four or all eight children. A homogeneous
root has score 1; one split into homogeneous children gives score 5 in 2D
or 9 in 3D.

Sort worlds by increasing \(C_h\), from the least to the most complex.
Equal scores receive the same rank. For two comparable worlds,
\[
 \W_a\preceq\W_b
 \quad\Longleftrightarrow\quad
 C_h(\W_a)\leq C_h(\W_b).
\]
Here, \(\preceq\) means ``has no greater spatial complexity under this
representation.'' Use the same dimension, physical domain, pitch, root
placement and orientation, and padding convention for the comparison.
For arrays that do not fit a dyadic square or cube, our implementation pads
only the high-index sides with blocked entries to the smallest enclosing
power-of-two square or cube. Padding and resolution affect the score, so
2D and 3D worlds are ranked separately.

\subsection{Example: complexity in POCGAGP}
\label{app:pocgagp-main}
\label{app:pocgagp}
\label{app:pocgagp-scaling}
The Partially Observable Cooperative Guard Art Gallery Problem (POCGAGP)
asks mobile agents to explore an initially unknown world and construct a
connected coverage network~\cite{pocgagp}. Boundary arrangement affects
occlusion and the positions needed for coverage and communication. For
this task, physical size or corner count alone does not describe all of
the relevant environmental variation.

Figure~\ref{fig:complexity-counterexamples} illustrates the visibility
component. Under unlimited omnidirectional line of sight, the large open
room can be seen from one observer, whereas the smaller maze contains
occluding walls and has a larger quadtree. The stepped room has more
corners than the maze, yet its entire interior can also be seen from one
observer. These are constructed visibility examples, not POCGAGP policy
simulations; connected exploration additionally depends on deployment,
movement and communication. Size can matter with finite sensor range,
and corners can matter through their arrangement. The point is that
neither quantity alone fixes the task's complexity.

\atlasfigure[.33\textheight]{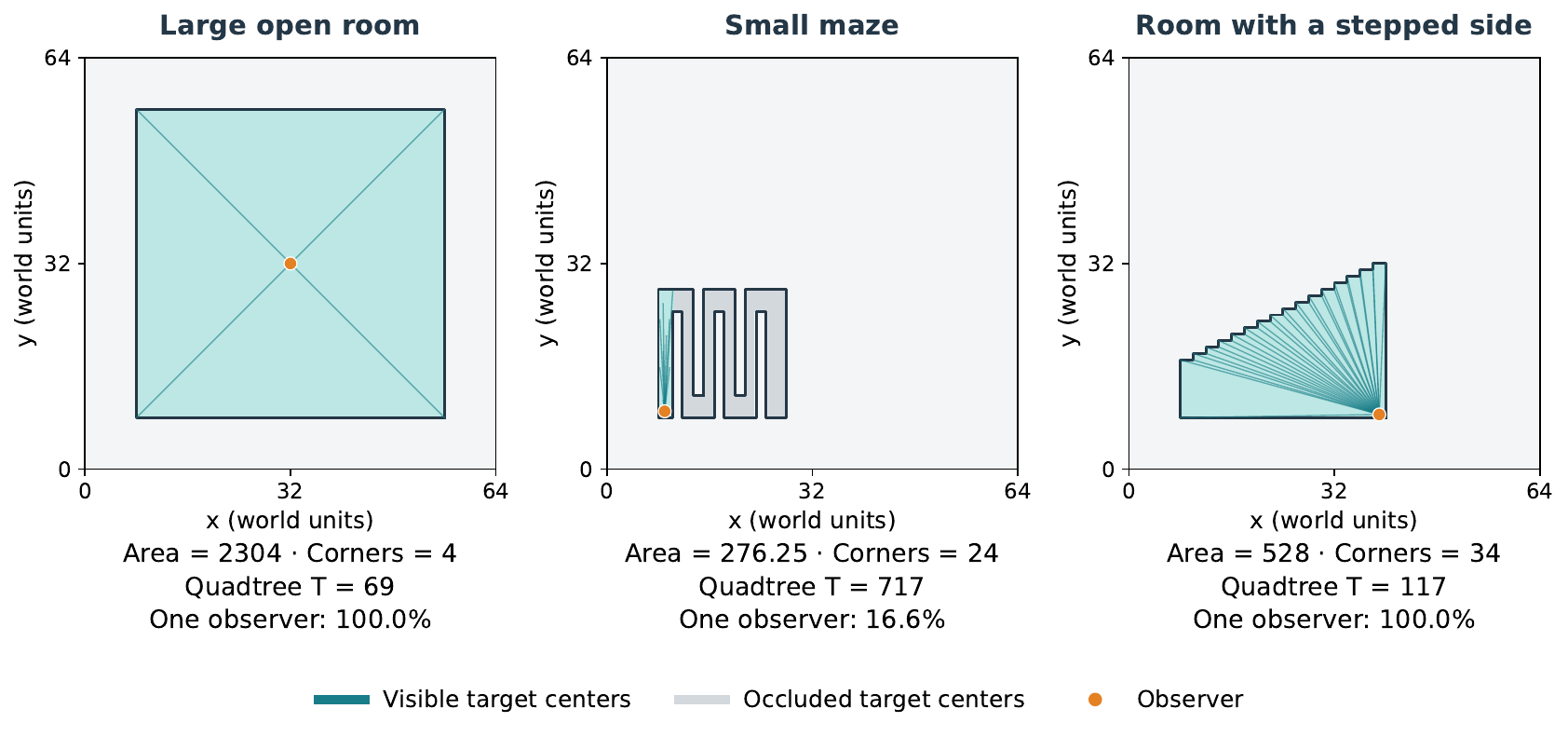}{Illustrative visibility cases for the POCGAGP complexity discussion. Size
and corner count alone do not determine coverage demand. All three
worlds are shown at the same physical scale and scored with the same
\(64\times64\) root and \(h=0.5\). The small maze has less free area
than the large open room, but a larger quadtree \((717>69)\). The
stepped room has more corners than the maze \((34>24)\), yet one observer
sees its entire interior. Orange marks the observer; colored cells show
visibility of their centers against continuous walls. Percentages use each
world's own finite target set. Section~\ref{app:complexity-counterexamples}
gives the construction and the whole-interior visibility argument.
\label{fig:complexity-counterexamples}}

The quadtree scores order the spatial detail of these masks, while the
visibility comparison shows why this ordering should not be interpreted
as an exact observer-count formula: the stepped room has a nontrivial
tree but is visible from one position.

\subsubsection{Observed relationship between tree size and deployment outcomes}
\begin{figure}[!htbp]
\centering
\begin{minipage}{.48\linewidth}
\centering
\includegraphics[trim=0 360bp 0 0,clip,width=\linewidth]{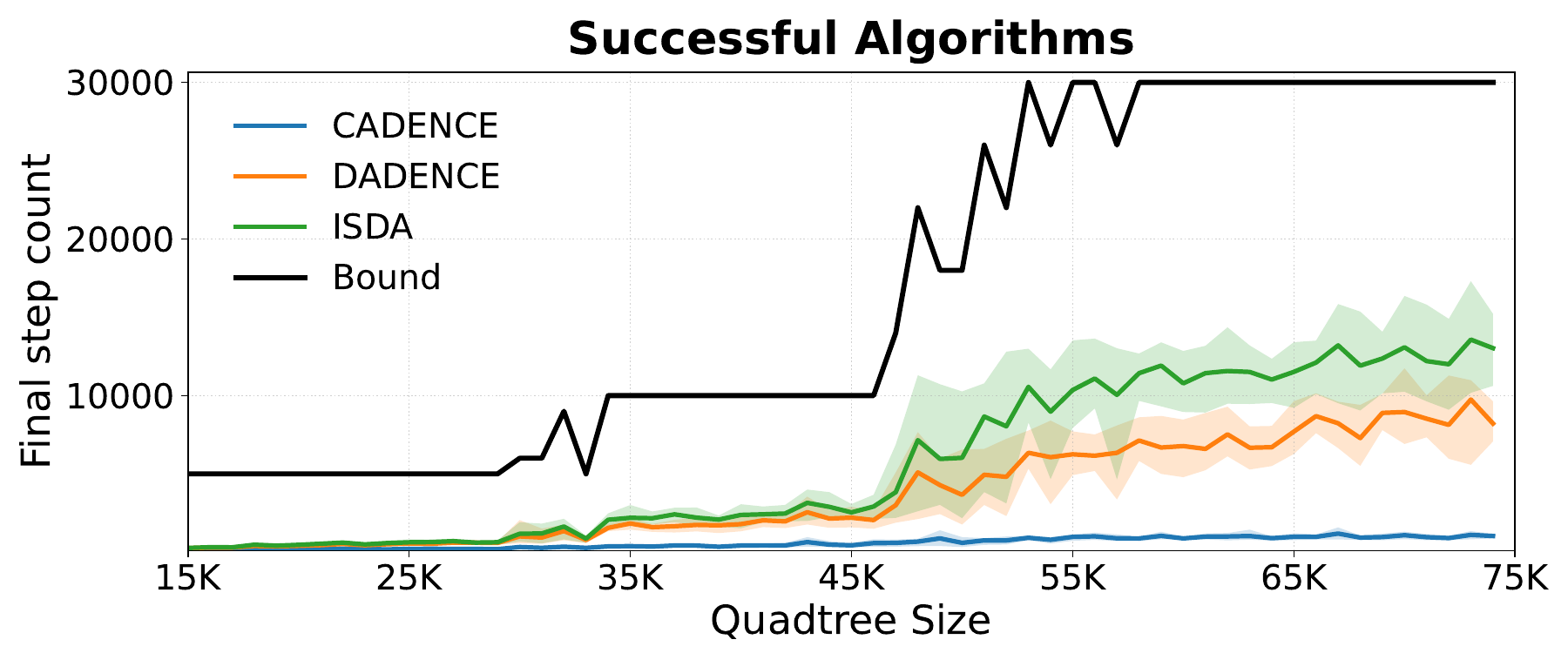}
\par{\small (a) Completion steps.\par}
\end{minipage}\hfill
\begin{minipage}{.48\linewidth}
\centering
\includegraphics[trim=0 360bp 0 0,clip,width=\linewidth]{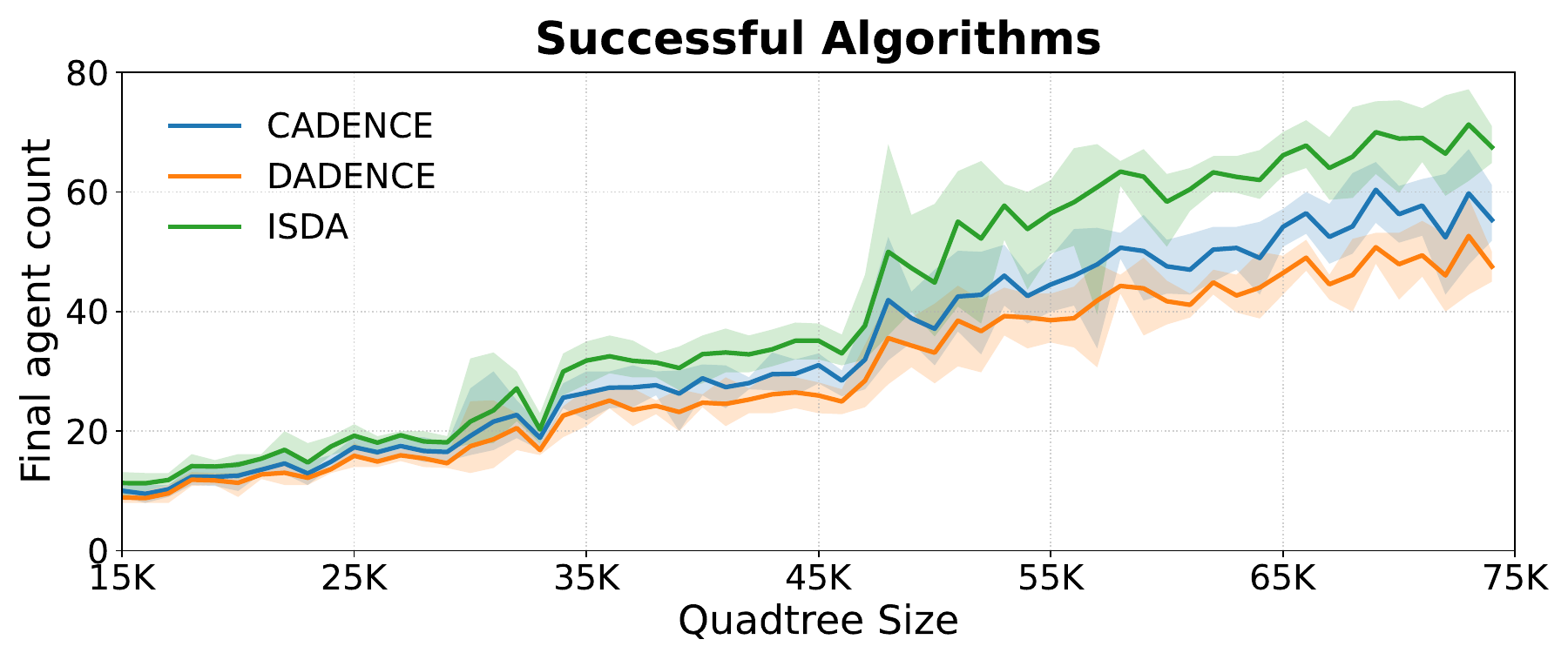}
\par{\small (b) Final agent population.\par}
\end{minipage}
\par\vspace{5pt}
\includegraphics[trim=0 360bp 0 0,clip,width=.48\linewidth]{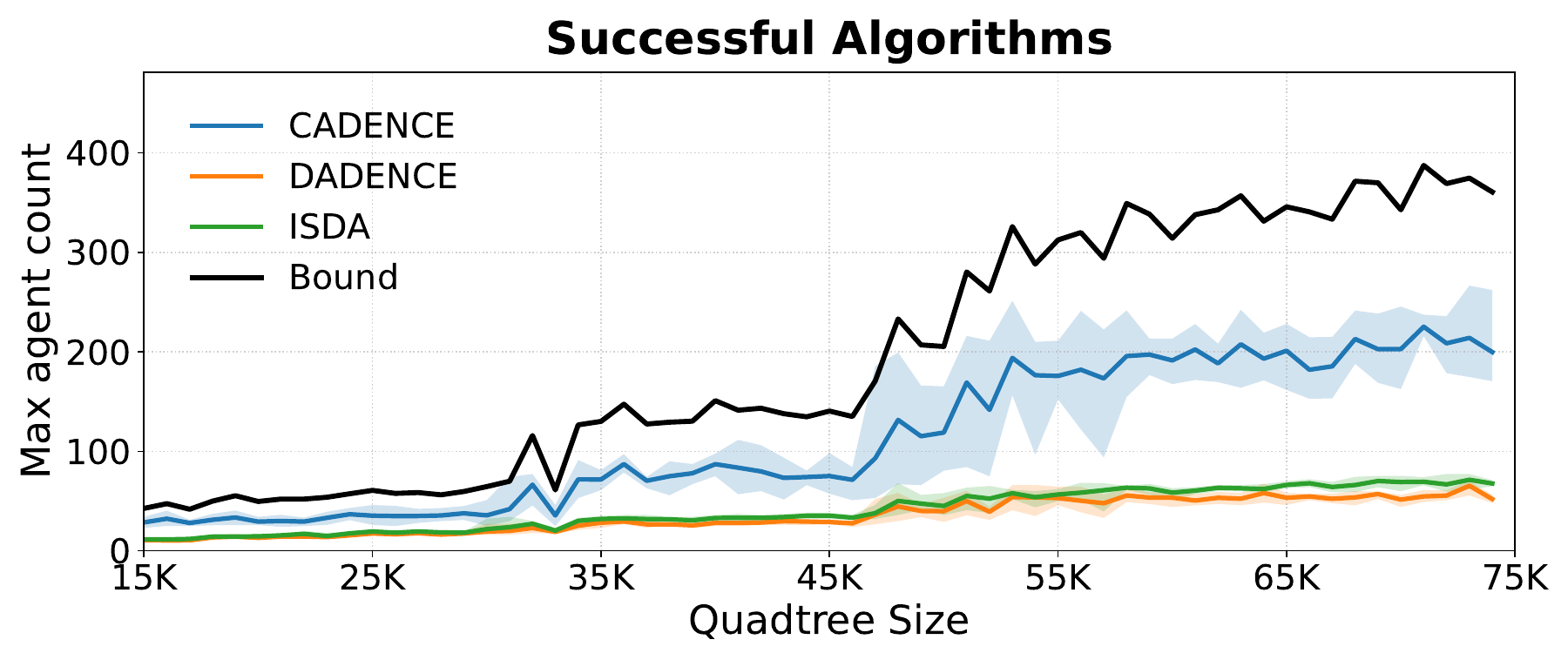}
\par{\small (c) Peak agent population.\par}
\caption{Successful-method panels from Figures 16--18 of
Meriaux et al.~\cite{pocgagp}. Original vector figure files are retained;
only the unsuccessful-method panels are cropped. The plotted values and
shaded regions are not recomputed or fitted here. Provenance records the
versioned source and file hashes.}
\label{fig:pocgagp-scaling}
\end{figure}

Figure~\ref{fig:pocgagp-scaling} gives measured POCGAGP evidence from
Meriaux et al.~\cite{pocgagp}. Completion steps, final agents and peak
agents are reported separately because methods respond differently to
increasing tree size.

Across the plotted complexity intervals, final agent populations rise for
all three methods. Completion-step growth is much stronger for DADENCE and
ISDA than for CADENCE; CADENCE instead has the larger peak population.
This provides empirical support for the descriptor defined in Section~\ref{sec:complexity-ranking}:
quadtree size captures environmental variation associated with agent and
completion-step demand in this cohort. The different responses motivate analyzing these
outcomes separately and testing the descriptor in non-orthogonal and 3D
families. They do not establish a universal scaling exponent.


\FloatBarrier

\subsection{Scores for the saved examples}
The native 2D masks have \(100\times100\) cells at pitch \(h=0.5\),
padded to a \(128\times128\) square. The native 3D masks have
\(70\times70\times42\) voxels at isotropic pitch \(h=50/70\),
padded to a \(128\times128\times128\) cube. Within each dimension,
the showcased worlds share the domain and root conventions of Section~6.
The measured total node counts are:
\hierarchyNativeTable
Examples 1, 2 and 3 use seeds 42, 53 and 64, respectively. For instance,
the three 2D curvilinear worlds rank as Example 1 \((1065)\),
Example 3 \((1229)\), then Example 2 \((1329)\). Thus generation order
does not determine complexity order. Counts and depth profiles are saved
in \path{results/hierarchy/showcase/}; \code{hierarchy.py} implements the
mask-based tree counter.

Figure~\ref{fig:hierarchy-native} shows the subdivisions for Example 1
of each class, using the same native masks and scores as the table.
\atlasfigure[.43\textheight]{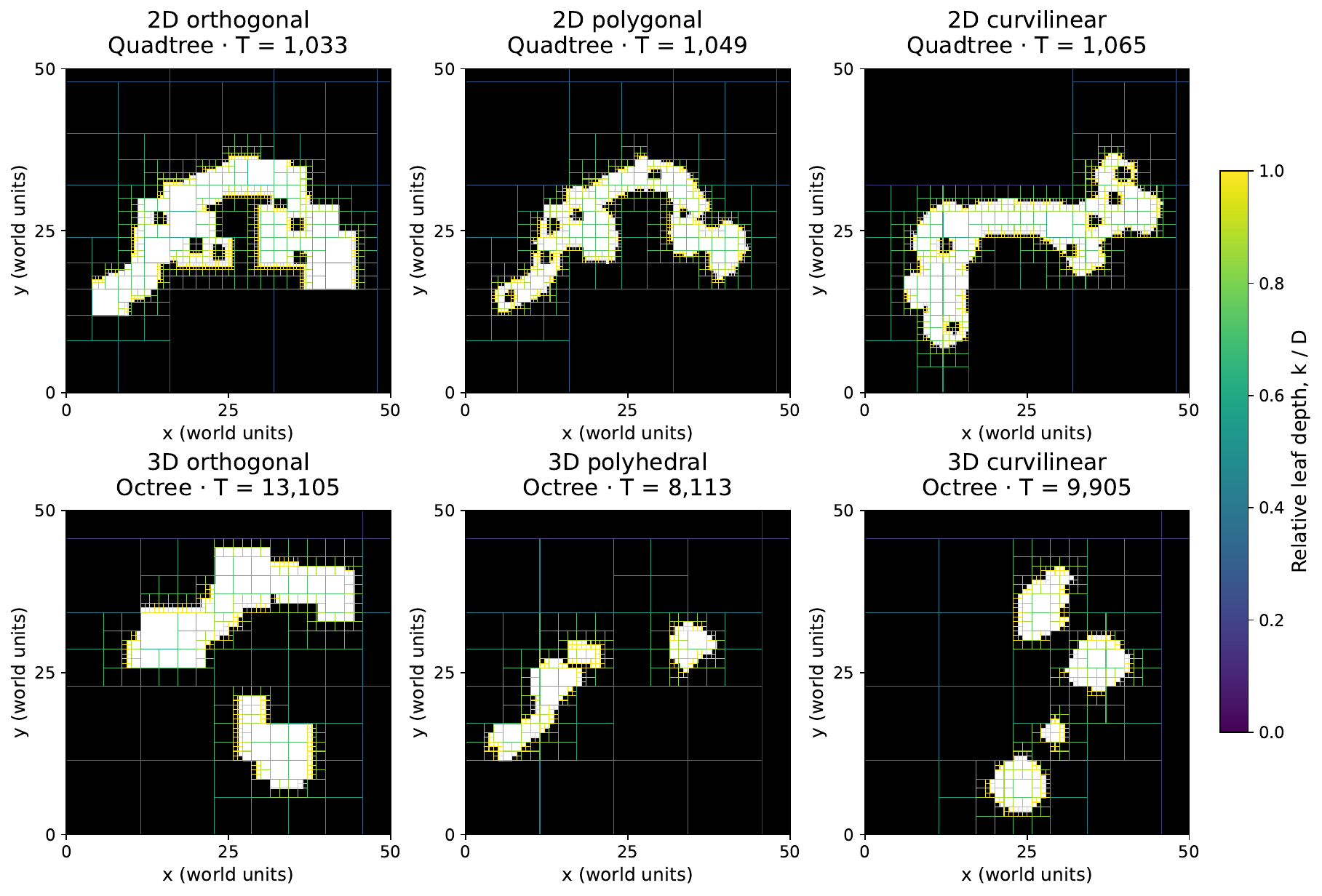}{Native-mask
quadtrees (top) and horizontal sections through native-mask octrees at
\(z=15\) (bottom). White indicates free entries and black indicates blocked
entries; colored outlines show retained leaves, with color indicating
relative subdivision depth. Large homogeneous regions remain coarse;
regions near boundaries subdivide more finely. Each \(T\) counts the
complete tree, including blocked padding outside the displayed world domain.
The bottom panels are sections of 3D octree leaves, not separate quadtrees.
\label{fig:hierarchy-native}}

\subsection{Construction of the illustrative worlds}
\label{app:complexity-counterexamples}
Figure~\ref{fig:complexity-counterexamples} uses three deterministic polygons
constructed for this comparison, rather than random generator outputs.
All three use a \(64\times64\) root, origin \((0,0)\), pitch \(h=0.5\),
whole-square containment and the same quadtree counting rule. Thus the
smaller maze's higher score is measured under the same representation
convention as the larger room's lower score.

The large room is the square \([8,56]\times[8,56]\).
The small maze starts with \([8,28]\times[8,28]\) and removes five
alternating wall strips of width 1.5, leaving gaps of width 3.5 at the
top or bottom. Its smaller free area \((276.25<2304)\) and smaller
bounding box coexist with a larger tree \((717>69)\).

The stepped room has 16 treads of width 2, with successive heights
9 through 24, translated by \((8,8)\). Its boundary has 34 corners,
more than the maze's 24, yet one observer at \((39,8.5)\) sees its
entire interior with unlimited omnidirectional line of sight. To see why,
use local coordinates relative to \((8,8)\), so the observer is
\((31,0.5)\). For a point to its left and above it, the sight segment
moves rightward and downward toward the observer, while the stepped
ceiling never decreases. Points below the observer remain below the
lowest ceiling. Points to its right share the final tread. In every
case the segment stays inside the room. This establishes visibility
of the whole interior, independently of target sampling.

For numerical replay, targets are all free-cell centers, checked against
continuous polygon walls. The Rust results agree with independent segment
containment checks for every target: the large room and stepped room have
100\% target coverage; the selected maze observer has 16.6\%.
A separate triangle-fan check verifies whole-region visibility for the
large and stepped rooms. These percentages refer to the stated observer
positions; no optimal observer count is inferred for the maze.
Saved polygons, occupancy masks, hierarchy profiles, observers, targets
and Rust query inputs and outputs are in
\path{datasets/complexity_counterexamples/}. The construction script is
\path{scripts/complexity_counterexamples.py}; use a fresh output directory
to reproduce the cases without replacing saved evidence.

\subsection{Historical metric reproduction and the 3D extension}

\label{app:pocgagp-reproduction}
\subsection*{Historical evidence and measurement contract}
The published panels in Figure~\ref{fig:pocgagp-scaling} use a selected cohort of
300 orthogonal worlds with five starts each. Worlds are grouped in
1,000-node intervals. The paper reports upward outcome trends within
fixed-size cohorts as well as across the complete collection~\cite{pocgagp}.
Black curves are resource budgets. Agent populations include the fixed
deployment pseudo-agent; separate exploratory analyses that exclude it
must not be combined without an explicit count conversion.

Recovered historical scripts resize grayscale maps to \(1024^2\), threshold
at 128 and count all nodes of a full-child region quadtree. Selection records
retain actual scores and their bins. This historical preprocessing differs
from the native-mask measurement convention in
Section~\ref{sec:complexity-ranking}. Numerical scores from these contracts
are not interchangeable.

\subsection*{Exact reproduction of the original scoring inputs}
The original scores preceded a corner-contact cleanup. The recovered
\code{image\_checker.py} changes a blocked cell to free when it encounters
a diagonal pair of free cells. The three copied experiment PNGs retain these
changes; they are preserved unchanged here.

We reconstruct the scoring inputs from the unchanged archived generator,
using its seed of 1 and one random stream across the successive batches of
5,000 \(50\times50\) maps and 5,000 \(100\times100\) maps, followed by the
\(250\times250\) batch. Replaying through the selected indices recovers
maps 685, 1815 and 1006 respectively. The pre-cleanup maps are saved
separately, resized to \(1024\times1024\) by bicubic interpolation,
thresholded at \(>128\), and counted. All three stored scores
reproduce exactly:
\hierarchyLegacyTable

The copied \(50\times50\) reference is unchanged by cleanup. The
\(100\times100\) reference differs from its regenerated scoring input by
one cell, and the \(250\times250\) reference differs by three cells. Applying
the archived cleanup rule to each regenerated map reproduces the entire
copied binary map exactly. Counting the copied maps instead gives
17,825, 43,669 and 72,537 nodes; the earlier discrepancies of \(-88\) and
\(-40\) are explained by the map modification.

The independent new counter and the recovered recursive counter agree on
all three resized pre-cleanup arrays. The package retains the regenerated
PNGs, resized Boolean masks, SHA-256 hashes and changed-cell coordinates.
To regenerate the inputs and require exact agreement with the original scores:
\begin{quote}\small
\texttt{python scripts/reproduce\_legacy\_references.py --verify}
\end{quote}
These checks establish reproduction for the three references, without
certifying every historical world's lineage or replaying the deployment
experiments.

The generator, scoring and cleanup scripts are preserved unchanged with
hashes. The replay extracts required generator functions without executing
archival module-level side effects. Original vector figures and versioned
arXiv sources remain preserved; their curves and shaded bands are not refitted.


\subsection*{A testable volumetric application}
The measured octrees establish a reproducible spatial descriptor for the
new 3D worlds. Testing its relationship to connected deployment requires
a 3D motion and sensing model, declared deployment starts, policy
implementations and termination conditions. Vertical motion, shafts,
sensor directionality and connectivity across elevations can introduce
effects beyond the node count.

Use a common physical pitch and root convention, retain physical extent
and geometry as covariates, and compare completion steps and agent
populations separately. Use the ranking convention in Section~\ref{sec:complexity-ranking}.
Keep repeated starts of a world together when estimating uncertainty
or testing on held-out worlds.
The 2D result motivates this experiment; a 3D deployment relationship
has not been measured here.



\end{document}